\documentclass[10pt,journal,compsoc]{IEEEtran}
\ifCLASSOPTIONcompsoc
\usepackage[nocompress]{cite}
\else
\usepackage{cite}
\fi
\ifCLASSINFOpdf
\usepackage[pdftex]{graphicx}
\usepackage{multirow}
\usepackage{amsmath}
\usepackage{amssymb}
\usepackage{booktabs}
\usepackage{caption}

\usepackage{xcolor} 
\usepackage{color} 
\usepackage{verbatim} 
\usepackage{diagbox}
\usepackage[pagebackref=true,breaklinks=true,letterpaper=true,colorlinks,citecolor=blue,linkcolor=blue,bookmarks=false]{hyperref}
\else

\fi
\usepackage{graphicx}  
\usepackage{float}
\usepackage{subfigure}
\usepackage{makecell}

\usepackage{colortbl}
\definecolor{yyellow}{rgb}{1, 1, 0.7}
\definecolor{oorange}{rgb}{1, 0.85, 0.7}
\definecolor{rred}{rgb}{1, 0.7, 0.7}
\newcommand{\best}{\cellcolor{rred}}
\newcommand{\sbest}{\cellcolor{oorange}}
\newcommand{\tbest}{\cellcolor{yyellow}}

\begin{document}
\title{GPS-Gaussian+: Generalizable Pixel-wise 3D Gaussian Splatting for Real-Time Human-Scene Rendering from Sparse Views}

\author{Boyao Zhou$^*$, Shunyuan Zheng$^*$, Hanzhang Tu, Ruizhi Shao,
        Boning Liu, Shengping Zhang,\\ Liqiang Nie and~Yebin Liu
\\{\ttfamily \url{https://yaourtb.github.io/GPS-Gaussian+}}
\thanks{$^*$ indicates equal contribution.}
\thanks{Boyao Zhou, Hanzhang Tu, Ruizhi Shao, Boning Liu and Yebin Liu are with Department of Automation, Tsinghua University, Beijing 100084, P.R.China.}
\thanks{Shunyuan Zheng and Shengping Zhang are with the School of Computer Science and Technology, Harbin Institute of Technology, Weihai 264209, P.R.China. Liqiang Nie is with the School of Computer Science and Technology, Harbin Institute of Technology, Shenzhen 518055, P.R.China.}
\thanks{Corresponding author: Shengping Zhang (s.zhang@hit.edu.cn).}
    }

\newcommand{\mname}{GPS-Gaussian}
\newcommand{\zhou}{\color{red}}
\markboth{Journal of \LaTeX\ Class Files,~Vol.~XX, No.~XX, XX~2024}%
{Shell \MakeLowercase{\textit{et al.}}: A Sample Article Using IEEEtran.cls for IEEE Journals}

\IEEEtitleabstractindextext{

\begin{abstract}
Differentiable rendering techniques have recently shown promising results for free-viewpoint video synthesis of characters. 
However, such methods, either Gaussian Splatting or neural implicit rendering, typically necessitate per-subject optimization which does not meet the requirement of real-time rendering in an interactive application.
We propose a generalizable Gaussian Splatting approach for high-resolution image rendering under a sparse-view camera setting.
To this end, we introduce Gaussian parameter maps defined on the source views and directly regress Gaussian properties for instant novel view synthesis without any fine-tuning or optimization.
We train our Gaussian parameter regression module on human-only data or human-scene data, jointly with a depth estimation module to lift 2D parameter maps to 3D space.
The proposed framework is fully differentiable with both depth and rendering supervision or with only rendering supervision.
We further introduce a regularization term and an epipolar attention mechanism to preserve geometry consistency between two source views, especially when neglecting depth supervision.
Experiments on several datasets demonstrate that our method outperforms state-of-the-art methods while achieving an exceeding rendering speed.
\end{abstract}

\begin{IEEEkeywords}
    3D Gaussian Splatting, Novel View Synthesis, Free Viewpoint Video
\end{IEEEkeywords}}
\maketitle
\IEEEdisplaynontitleabstractindextext
\IEEEpeerreviewmaketitle

\IEEEraisesectionheading{\section{Introduction}\label{sec:introduction}}
\IEEEPARstart{F}{ree}-Viewpoint Video (FVV) synthesis from sparse input views is a challenging and crucial task in computer vision, which is largely used in sports broadcasting, stage performance and telepresence systems~\cite {lawrence2021project, tu2024tele}. However, early attempts~\cite{chen1993view, oh2001image} try to solve this problem through a weighted blending mechanism~\cite{wilburn2005high} by using a huge number of cameras, which dramatically increases computational cost and latency. On the other hand, NeRF-like differentiable volumetric rendering techniques~\cite{mildenhall2020nerf, peng2021neural-body, muller2022instant-ngp, fridovich2022plenoxels} can synthesize novel views under sparse camera setting~\cite{shao2023tensor4d}, but typically suffer from per-scene optimization~\cite{mildenhall2020nerf, peng2021neural-body, muller2022instant-ngp, fridovich2022plenoxels}, slow rendering speed~\cite{mildenhall2020nerf, peng2021neural-body} and overfitting to input views~\cite{yu2021pixelnerf}. 

\begin{figure}[htpb]
\vspace{-0.6cm}
\includegraphics[width=\linewidth]{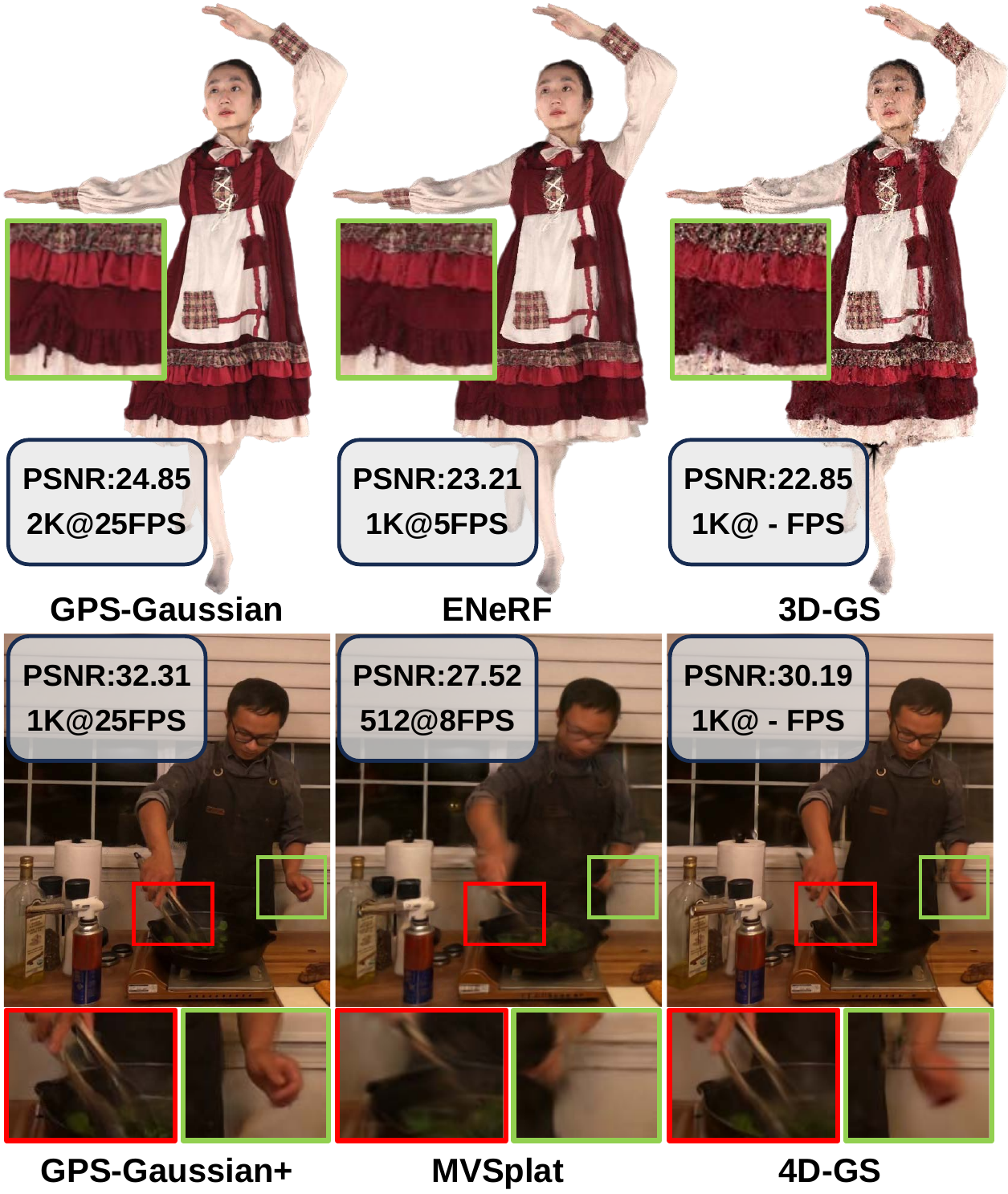}
\vspace{-0.6cm}
\caption{\textbf{High-fidelity and real-time rendering.} On the top, GPS-Gaussian produces $2K$-resolution rendering of character, while GPS-Gaussian+ renders novel views of human-centered scenes on the bottom. Our methods outperform the state-of-the-art feed-forward implicit rendering method ENeRF~\cite{lin2022enerf}, explicit rendering method MVSplat~\cite{chen2024mvsplat} and optimization-based methods 3D-GS~\cite{kerbl2023_3dgs} and 4D-GS~\cite{Wu2024_4dgs}.}
\vspace{-0.6cm}
\label{fig:teaser}
\end{figure}

In contrast, point-based rendering~\cite{levoy1985use, zwicker2001surface, wiles2020synsin, lassner2021pulsar} has drawn long-lasting attention thanks to its high-speed, and even real-time, rendering performance. Once integrated with neural networks, point-based graphics~\cite{aliev2020npbg, rakhimov2022npbg++} realize a promising explicit representation with comparable realism and extremely superior efficiency in FVV tasks~\cite{aliev2020npbg, rakhimov2022npbg++}. Recently, 3D Gaussian Splatting (3D-GS)~\cite{kerbl2023_3dgs} introduces a new representation that the point clouds are formulated as 3D Gaussians with a series of learnable properties including 3D position, color, opacity and anisotropic covariance. By applying $\alpha$-blending~\cite{kopanas2022neural}, 3D-GS provides not only a more reasonable and accurate mechanism for back-propagating the gradients but also a real-time rendering efficiency for complex scenes. Despite realizing a real-time inference, Gaussian Splatting relies on per-scene~\cite{kerbl2023_3dgs} or per-frame~\cite{luiten2023dynamic} parameter optimization for several minutes. It is therefore impractical in interactive scenarios as it necessitates the re-optimization of Gaussian parameters once the character or the scene changes.

More recently, some generalizable Gaussian Splatting methods~\cite{szymanowicz2023splatter, charatan2023pixelsplat, chen2024mvsplat, liu2024mvsgaussian} have been proposed to explore novel-view synthesis in a feed-forward way. In general, such methods leverage a learnable geometry prior with a differentiable rendering pipeline to achieve feed-forward inference. For example, Splatter Image~\cite{szymanowicz2023splatter} regresses directly Gaussians' positions and other properties from a single image. In such an ill-posed setting, 3D consistency is hardly held and image quality is extremely low. PixelSplat~\cite{charatan2023pixelsplat}, MVSplat~\cite{chen2024mvsplat} and MVSGaussian~\cite{liu2024mvsgaussian} utilize probabilistic depth estimation and multiplane sweeping to represent geometry from multiple source view images. Although such geometry cues allow 3D-GS to generalize to some static scenes, the probabilistic representations can easily make floating artifacts due to uncertain Gaussian positions. Moreover, they dramatically increase the computational cost for inference and no longer render novel view image in real-time, even with a low resolution of $256 \times 256$. 

In this paper, we propose to integrate binocular stereo-matching~\cite{li2022practical, weinzaepfel2023croco, lipson2021raft-stereo} as a geometry cue with 3D-GS rendering pipeline to achieve a generalizable Gaussian Splatting. Given a pair of images, stereo-matching can determinately calculate the disparity by searching the cost volume built on the extracted features of two source views. Based on the epipolar geometry, it is straightforward to transform disparity into depth given camera parameters. This determinant geometry representation can be better supervised with rendering loss than probabilistic ones because Gaussian Splatting necessitates a certain position of each primitive.

Specifically, we introduce 2D Gaussian parameter (depth residual, color, scaling, rotation, opacity) maps which are defined on source view image planes, instead of unstructured point clouds. These Gaussian parameter maps allow us to represent a scene with pixel-wise parameters, \textit{i.e.} each pixel corresponding to a specific Gaussian point. Additionally, it enables the application of efficient 2D convolution networks rather than expensive 3D operators. Given the estimated depth map via binocular stereo-matching, it is efficient to lift 2D parameter maps to 3D Gaussian points. Such unprojected Gaussian points from both the two source views constitute the representation of scene and novel view images can be rendered with splatting technique~\cite{kerbl2023_3dgs}. Since the whole pipeline is fully differentiable, we can jointly train an iterative stereo matching-based depth estimation~\cite{lipson2021raft-stereo} along with our Gaussian parameter regression with both depth and rendering loss or with only rendering loss.

A preliminary version of this work has been published as a highlight paper~\cite{zheng2024gpsgaussian} in CVPR 2024, in which we propose a real-time framework of human novel view synthesis and train this framework on synthetic human-only data with depth and rendering loss.
In the current version, we aim to extend it to human-scene scenarios and no longer require matting technique or depth supervision. Note that wrong matting leads to floating artifacts and it is not trivial to acquire high-quality geometry or depth information from complex human-centered scene data.
To achieve high-quality rendering without depth supervision, an epipolar attention is applied in the shared feature extraction module (Sec.~\ref{sec:feature_extraction}), which can improve stereo-matching accuracy and rendering consistency.
In addition, a depth residual map (Sec.~\ref{sec:position_res}) is introduced in order to recover high-frequency details from the predicted depth of stereo-matching when lacking depth supervision. 
Furthermore, we propose a regularization term in Sec.~\ref{sec:geo_regular} to preserve geometry consistency between the two source views and improve the overall stability of the training process when lacking the ground truth of depth.
Thanks to these adaptive components, our network can be trained with only rendering loss, making it scalable for more general human-scene scenarios.

In practice, we are able to synthesize high-fidelity free-viewpoint video around 25 FPS on a single modern graphics card. Leveraging the rapid rendering capabilities and broad generalizability inherent in our proposed method, an unseen character with or without background can be instantly rendered without necessitating any fine-tuning or optimization. In summary, our contributions can be summarized as:
\noindent
\begin{itemize}
    \item We introduce a generalizable 3D Gaussian Splatting methodology that employs pixel-wise Gaussian parameter maps defined on 2D source image planes to formulate 3D Gaussians in a feed-forward manner.

    \item We propose a fully differentiable framework composed of an iterative depth estimation module and a Gaussian parameter regression module. The intermediate depth prediction bridges the two components and allows them to benefit from joint training.

    \item We introduce a regularization term and an epipolar attention mechanism to preserve geometry consistency between the two source views when using only rendering loss. Our method generalizes well to unseen characters even in complicated scenes.
    
    \item We develop a real-time FVV system that achieves high-resolution rendering of characters in the scene without any geometry supervision.

\end{itemize}

\section{Related Work}

\noindent\textbf{Neural Implicit Representation.}
Neural implicit function has recently aroused a surge of interest to represent complicated scenes, in form of occupancy fields~\cite{mescheder2019occupancy, saito2019pifu, saito2020pifuhd, hong2021stereopifu}, radiance fields~\cite{mildenhall2020nerf, peng2021neural-body, fang2022TiNeuVox, zhao2022humannerf, weng2022humannerf, guo2023vid2avatar} and signed distance functions~\cite{park2019deepsdf, wang2021neus, wang2023neus2, shao2023tensor4d, zhou2023human}.
Implicit representation shows the advantage in memory efficiency and topological flexibility for human representation~\cite{hong2021stereopifu, zheng2021pamir, xiu2022icon} or scene reconstruction~\cite{peng2020convolutional,chibane2020ndf}, especially in a pixel-aligned feature query manner~\cite{saito2019pifu, saito2020pifuhd}.
However, each queried point is processed through the full network, which dramatically increases computational complexity.  
More recently, numerous methods have extended Neural Radiance Fields (NeRF)~\cite{mildenhall2020nerf} to static human modeling~\cite{shao2022doublefield, chen2023gm-nerf} and dynamic human modeling from sparse multi-view cameras~\cite{peng2021neural-body, zhao2022humannerf, shao2023tensor4d} or a monocular camera~\cite{weng2022humannerf, jiang2022neuman, guo2023vid2avatar}.
However, these methods typically require a per-subject optimization process and it is non-trivial to generalize these methods to unseen subjects. Previous attempts, \textit{e.g.}, PixelNeRF~\cite{yu2021pixelnerf}, IBRNet~\cite{wang2021ibrnet}, MVSNeRF~\cite{chen2021mvsnerf} and ENeRF~\cite{lin2022enerf} resort to image-based features as potent priors for feed-forward scene modeling.
Despite the great progress in accelerating the scene-specific NeRF~\cite{yu2021plenoctrees, fridovich2022plenoxels, muller2022instant-ngp, li2023nerfacc}, efficient generalizable NeRF for interactive scenarios remains to be further elucidated.

\noindent\textbf{Deep Image-based Rendering.}
Image-based rendering, or IBR in short, synthesizes novel views from a set of multi-view images with a weighted blending mechanism, which is typically computed from a geometry proxy.
\cite{riegler2020free, riegler2021stable} deploy multi-view stereo from dense input views to produce mesh surfaces as a proxy for image warping.
DNR~\cite{thies2019dnr} directly produces learnable features on the surface of mesh proxies for neural rendering.
Obtaining these proxies is not straightforward since high-quality multi-view stereo and surface reconstruction requires dense input views.
Point clouds from SfM~\cite{meshry2019neural, pittaluga2019revealing} or depth sensors~\cite{martin2018lookingood, nguyen2022hsv-net} can also be engaged as geometry proxies.
These methods highly depend on the performance of 3D reconstruction algorithms or the quality of depth sensors. 
FWD~\cite{cao2022fwd} designs a network to refine depth estimations, then explicitly warps pixels from source views to novel views with the refined depth maps.
FloRen~\cite{shao2022floren} utilizes a coarse human mesh reconstructed by PIFu~\cite{saito2019pifu} to render initialized depth maps for novel views.
Arguably FloRen~\cite{shao2022floren} is most related to our preliminary work GPS-Gaussian~\cite{zheng2024gpsgaussian}, as it also realizes $360^\circ$ free view human performance rendering in real-time.
However, the appearance flow in FloRen merely works in 2D domains, where the rich geometry cues and multi-view geometric constraints only serve as 2D supervisions.
The difference is that our approach lifts 2D priors into 3D space and utilizes the point representation to synthesize novel views in a fully differentiable manner.

\noindent\textbf{Point-based Graphics.}
Point-based representation has shown great efficiency and simplicity for various 3D human-centered tasks~\cite{liu2009point, zhou2020reconstructing, ma2021power, lin2022eccv, zheng2023pointavatar, zhang2023closet, xu20244k4d}.
Previous attempts integrate point cloud representation with 2D neural rendering~\cite{aliev2020npbg, rakhimov2022npbg++} or NeRF-like volume rendering~\cite{xu2022point-nerf, su2023npc}.
Still, such a hybrid architecture does not exploit the rendering capability of point clouds and takes a long time to optimize on different scenes.
Then differentiable point-based~\cite{wiles2020synsin} and sphere-based~\cite{lassner2021pulsar, xu20244k4d} rendering have been developed, which demonstrates promising rendering qualities, especially attaching them to a conventional network pipeline~\cite{cao2022fwd, nguyen2022hsv-net}.
In addition, isotropic points can be substituted by a more reasonable Gaussian point modeling~\cite{kerbl2023_3dgs, luiten2023dynamic}, to realize a rapid differentiable rendering framework with a splatting technique. 
This advanced representation has showcased prominent performance in concurrent 3D human-centered work~\cite{hu2024gaussianavatar, shao2024control4d, xu2024gaussian, li2024animatable, liu2024humangaussian}.
However, a per-scene or per-subject optimization strategy limits its real-world application.
Although \cite{luiten2023dynamic, sun20243dgstream} accelerate partly the optimization process by using an on-the-fly strategy, they struggle to handle topology change in dynamic scenes.
In this paper, we go further to generalize 3D Gaussians across diverse subjects while maintaining its fast and high-quality rendering properties.

\noindent\textbf{Free-Viewpoint Video.}
Targeting different applications, there are two feasible schemes to produce free-viewpoint videos, one uses a compact 4D representation~\cite{li2022neural, shao2023tensor4d, attal2023hyperreel, cao2023hexplane, li2024spacetime, Wu2024_4dgs}, and the other formulate an individual 3D representation for each discrete timestamp, which can be further subdivided into on-the-fly optimization methods~\cite{luiten2023dynamic, sun20243dgstream, xu20244k4d, jiang2024hifi4g} and feed-forward inference methods~\cite{suo2021neuralhumanfvv, zheng2024gpsgaussian, kwon2024generalizable, liu2024mvsgaussian}.
The 4D representations~\cite{attal2023hyperreel, cao2023hexplane, Wu2024_4dgs, li2024spacetime} cater to volumetric video, which can be played back and viewed from any viewpoint at any time, but the performance degrades when the capturing time goes longer.
On the contrary, the on-the-fly optimization method~\cite{jiang2024hifi4g} excels at handling long-time sequences.
They can realize similar experiences after applying customized compression designs but they typically have higher memory costs than 4D methods.
Nevertheless, despite having been accelerated, the essential optimization process is far from real-time.
Thus, we orient towards feed-forward methods for interactive scenarios.
Among them, MonoFVV~\cite{guo2017real} and Function4D~\cite{yu2021function4d} implement RGBD fusion with depth sensors to attain real-time human rendering.
The large variation in pose and clothing makes the feed-forward generalizable free-view rendering a more challenging task, thus recent work~\cite{kwon2021nhp, chen2023gm-nerf, pan2023transhuman, suo2021neuralhumanfvv, kwon2024generalizable} simplifies the problem by leveraging human priors~\cite{loper2015smpl, saito2019pifu}.
However, an inaccurate prior estimation would mislead the final result.
For more general dynamic scenarios, \cite{lin2022enerf} relies on expensive probabilistic geometry estimation, thus they can hardly achieve real-time free-viewpoint rendering, even integrated with Gaussian Splatting~\cite{charatan2023pixelsplat, chen2024mvsplat, liu2024mvsgaussian}.

\begin{figure*}
  \centering
  \includegraphics[width=\textwidth]{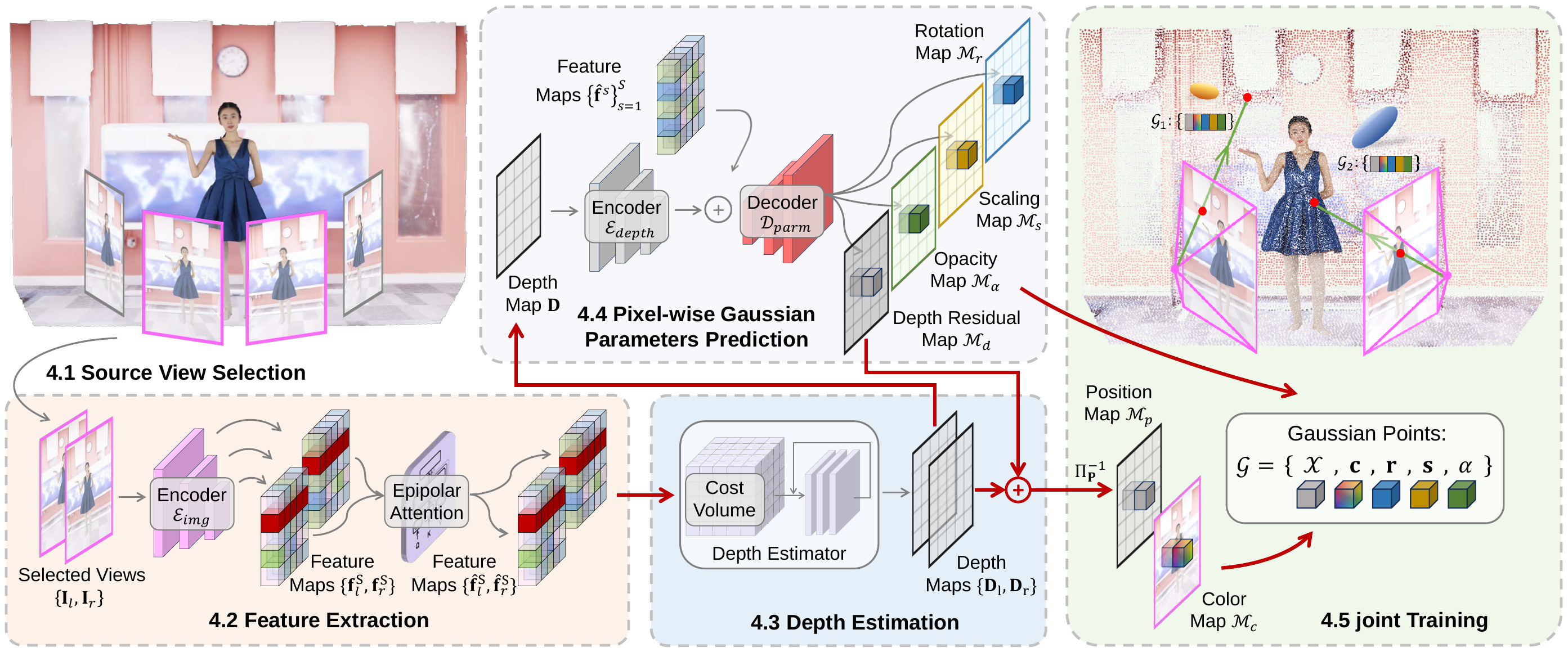}
  \caption{\textbf{Overview of GPS-Gaussian+.} Given RGB images of a human-centered scene with sparse camera views and a target novel viewpoint, we select the adjacent two views on which to formulate our pixel-wise Gaussian representation. We extract the image features by using epipolar attention and then conduct an iterative depth estimation. For each source view, the RGB image serves as a color map, while the other parameters of 3D Gaussians are predicted in a pixel-wise manner. The Gaussian parameter maps defined on 2D image planes of both views are further unprojected to 3D space via refined depth maps and aggregated for novel view rendering. The fully differentiable framework enables a joint training mechanism with only rendering loss and geometry regularization.}
  \label{fig:pipeline}
\end{figure*}

\section{Preliminary}
\label{sec:preliminary} 
Since the proposed GPS-Gaussian+ harnesses the power of 3D-GS~\cite{kerbl2023_3dgs}, we give a brief introduction in this section.

3D-GS models a static 3D scene explicitly with point primitives, each of which is parameterized as a scaled Gaussian with 3D covariance matrix $\mathbf{\Sigma}$ and mean $\mathbf{\mu}$
\begin{equation}
\label{formula:gaussian formula}
    G(\mathcal{X})~= e^{-\frac{1}{2}(\mathcal{X}-\mu)^{T}{\mathbf{\Sigma}}^{-1}(\mathcal{X}-\mu)}
\end{equation}
In order to be effectively optimized by gradient descent, the covariance matrix $\mathbf{\Sigma}$ can be decomposed into a scaling matrix $\mathbf{S}$ and a rotation matrix $\mathbf{R}$ as
\begin{equation}
\label{formula:covariance decomposition}
    \mathbf{\Sigma} = \mathbf{RSS}^T\mathbf{R}^T
\end{equation}
Following~\cite{zwicker2002ewa}, the projection of Gaussians from 3D space to a 2D image plane is implemented by a view transformation $\mathbf{W}$ and the Jacobian of the affine approximation of the projective transformation $\mathbf{J}$. The covariance matrix ${\mathbf{\Sigma}}^{\prime}$ in 2D space can be computed as
\begin{equation}
    \mathbf{\Sigma^{\prime}} = \mathbf{JW\Sigma W}^T\mathbf{J}^T
\end{equation}
followed by a point-based alpha-blend rendering which bears similarities to that used in NeRF~\cite{mildenhall2020nerf}, formulated as
\begin{equation}
\label{formula: splatting&volume rendering}
    \mathbf{C}_{color} = \sum_{i\in N}\mathbf{c}_i \alpha_i \prod_{j=1}^{i-1} (1-\alpha_i)
\end{equation}
where $\mathbf{c}_i$ is the color of each point, and density $\alpha_i$ is reasoned by the multiplication of a 2D Gaussian with covariance ${\mathbf{\Sigma}}^{\prime}$ and a learned per-point opacity~\cite{yifan2019differentiable}. The color is defined by spherical harmonics (SH) coefficients in \cite{kerbl2023_3dgs}.

To summarize, the original 3D Gaussians methodology characterizes each Gaussian point by the following attributes: (1) a 3D position of each point $\mathcal{X} \in \mathbb{R}^3$, (2) a color defined by SH $\mathbf{c} \in \mathbb{R}^k$ (where $k$ is the freedom of SH basis), (3) a rotation parameterized by a quaternion $\mathbf{r} \in \mathbb{R}^4$, (4) a scaling factor $\mathbf{s} \in \mathbb{R}_+^3$, and (5) an opacity $\alpha \in [0, 1]$.

\section{Method}
\label{sec:method}
The overview of our method is illustrated in Fig.~\ref{fig:pipeline}.
Given RGB images of a human-centered scene with sparse camera views, our method aims to generate high-quality free-viewpoint video of the performer in real-time.
Once given a target novel viewpoint, we select the two neighboring views from sparse cameras (Sec.~\ref{sec:selection}). 
Then, image features are extracted from the two input images with a shared image encoder by using epipolar attention (Sec.~\ref{sec:feature_extraction}), and they are further used to predict the depth maps for both source views with a binocular depth estimator (Sec.~\ref{sec:depth_estimation}).
The colors of 3D Gaussians are directly determined by the corresponding source view pixels, while other parameters of 3D Gaussians are predicted in a pixel-wise manner when feeding the predicted depth values and the former image features into a network (Sec.~\ref{sec:gaussian_network}).
Combined with RGB map of the source view image, these parameter maps formulate the Gaussian representation in 2D image planes and are further unprojected to 3D space with the estimated depth. 
The unprojected Gaussians from both views are aggregated and rendered to the target viewpoint in a differentiable way, which allows for end-to-end training (Sec.~\ref{sec:render}), even with only rendering loss.

\subsection{Source View Selection}
\label{sec:selection}
As a binocular stereo method, we synthesize the target novel view with two adjacent source views.
Given $N$ input images $\{\mathbf{I}_n\}_{n=1}^N$, with their camera position $\{C_n\}_{n=1}^N$, source views can be represented by $\mathbf{V_n}=C_n-O$, where $O$ is the center of the scene.
Similarly, the target novel view rendering can be defined as $I_{tar}$ with camera position $C_{tar}$ and view $\mathbf{V_{tar}}=C_{tar}-O$.
By conducting a dot product of all source view vectors and the novel view vector, the nearest two views $(v_{l}, v_{r})$ can be selected as the `working set' of binocular stereo, where $l$ and $r$ stand for `left' and `right' view, respectively.

\subsection{Feature Extraction}
\label{sec:feature_extraction}
The selected images are encoded with a feature extraction module in order to search the corresponding features from one view to another.
Once two source view images are rectified, $\mathbf{I}_{l}, \mathbf{I}_{r}\in[0,1]^{H\times W\times 3}$ are fed to a shared image encoder $\mathcal{E}_{img}$ with several residual blocks and downsampling layers to extract dense feature maps $\mathbf{f}^s \in \mathbb{R}^{H/2^s\times W/2^s\times D_s}$ where $D_s$ is the dimension at the $s$-th feature scale
\begin{equation}
\label{formula:image encoder}
    \langle\{\mathbf{f}_l^s\}_{s=1}^S, \{\mathbf{f}_r^s\}_{s=1}^S\rangle = \mathcal{E}_{img}(\mathbf{I}_{l}, \mathbf{I}_{r})
\end{equation}
where we set $S=3$ in our experiments. 

Since the image encoder $ \mathcal{E}_{img}$ is independent of each other view, it struggles to extract informative features when lacking depth supervision.
Thus, we propose to conduct an epipolar attention module on the bottleneck features $\mathbf{f}_{l,r}^S$, in order to exchange useful information from each other view. Note that corresponding pixels from the two rectified images are located on the same horizontal epipolar line. In practice, we rearrange feature map into $H/2^S$ line features $\mathbf{f}^e \in \mathbb{R}^{W/2^S\times D_S}$ and employ multi-head attention~\cite{vaswani2017attention} $Att$ along each epipolar line
\begin{align}
\begin{split}
\label{formula:epipolar_att}
    \langle\mathbf{Q,K,V}\rangle &= \langle\mathbf{f}^e \mathbf{W}^Q, \mathbf{f}^e \mathbf{W}^K, \mathbf{f}^e \mathbf{W}^V \rangle \\
    \hat{\mathbf{f}}_i &= \mathbf{f}^e_i + Att(\mathbf{Q}_i, \mathbf{K}_j, \mathbf{V}_j)
\end{split}
\end{align}
where $\{i,j\} = \{l,r\}$ or $\{r,l\}$. Processed line features $\{\hat{\mathbf{f}}_k\}_{k=1}^{H/2^S}$ of source view are concatenated into feature map $\hat{\mathbf{f}}^S \in \mathbb{R}^{H/2^S \times W/2^S\times D_S}$ following some convolution operations. Such an attention-based encoder $ \mathcal{E}^{att}$ significantly increases the perceptive field of the feature extractor so that the extracted features can be used to build cost volume in the following section.

\subsection{Depth Estimation}
\label{sec:depth_estimation}
The depth map is the key component of our framework which lifts the 2D image planes to 3D Gaussian representation.
Note that, depth estimation in binocular stereo is equivalent to disparity estimation.
For each pixel coordinate $x = (u, v)$ in one view, disparity estimation $\phi_{disp}$ aims to find its corresponding coordinate $(u+\phi_{disp}(u), v)$ in another view, considering the displacement of each pixel is constrained to a horizontal line in rectified stereo.
Since there is a one-to-one mapping between disparity maps and depth maps given camera parameters, we do not distinguish them in the following sections.
Inspired by \cite{lipson2021raft-stereo}, we implement this module in an iterative manner mainly because it avoids using prohibitively slow 3D convolutions to filter the cost volume.
Given the processed feature maps $\hat{\mathbf{f}}_l^S, \hat{\mathbf{f}}_r^S$, we compute a 3D cost volume $\mathbf{C} \in \mathbb{R}^{H/2^S\times W/2^S\times W/2^S}$ using matrix multiplication
\begin{equation}
\label{formula:cost volume}
    \mathbf{C}(\hat{\mathbf{f}}_l^S, \hat{\mathbf{f}}_r^S), \quad C_{ijk} = \sum_h (\hat{\mathbf{f}}_l^S)_{ijh} \cdot (\hat{\mathbf{f}}_r^S)_{ikh}
\end{equation}
Then, an iterative update mechanism predicts a sequence of depth estimations $\{\mathbf{d}_l^t\}_{t=1}^T$ and $\{\mathbf{d}_r^t\}_{t=1}^T$ by looking up in volume $\mathbf{C}$, where $T$ is the update iterations.
For more details about the update operators, please refer to \cite{teed2020raft}.
The outputs of the final iterations ($\mathbf{d}_l^{T}$, $\mathbf{d}_r^{T}$) are upsampled to full image resolution via a convex upsampling.
The depth estimation module $\Phi_{depth}$ can be formulated as
\begin{equation}
\label{formula:disp estimation}
    \langle\mathbf{D}_l, \mathbf{D}_r\rangle = \Phi_{depth}(\hat{\mathbf{f}}_l^S, \hat{\mathbf{f}}_r^S, K_l, K_r)
\end{equation}
where $K_l$ and $K_r$ are the camera parameters, $\mathbf{D}_l, \mathbf{D}_r \in \mathbb{R}^{H \times W \times 1}$ are the depth estimations.
The classic binocular stereo methods estimate the depth for `reference view' only, while we pursue depth maps for both input views with a shared-weight network to serve as the position of Gaussian points, which results in a decent efficiency increase.

\subsection{Pixel-wise Gaussian Parameters Prediction}
\label{sec:gaussian_network}
In 3D-GS~\cite{kerbl2023_3dgs}, each Gaussian point in 3D space is characterized by attributes $\mathcal{G}=\{\mathcal{X}, \mathbf{c}, \mathbf{r}, \mathbf{s}, \alpha\}$, which represent 3D position, color, rotation, scaling and opacity, respectively.
In this section, we introduce a pixel-wise manner to formulate 3D Gaussians in 2D image planes.
Specifically, the proposed Gaussian maps $\mathbf{G}$ are defined as
\begin{equation}
\label{formula:gs_def}
    \mathbf{G}(x) = \{\mathcal{M}_p(x), \mathcal{M}_c(x), \mathcal{M}_r(x), \mathcal{M}_s(x), \mathcal{M}_\alpha(x)\}
\end{equation}
where $x$ is the coordinate of a valid pixel in a rectified image plane, $\mathcal{M}_p, \mathcal{M}_c, \mathcal{M}_r, \mathcal{M}_s, \mathcal{M}_\alpha$ represents Gaussian parameter maps of position, color, rotation, scaling and opacity, respectively.
\subsubsection{Color Map}
\label{sec:color_map}
Considering our human-centered scenario is predominantly characterized by diffuse reflection, instead of predicting the sphere harmonic (SH) coefficients, we directly use the source RGB image as the color map
\begin{equation}
\label{formula:gs_color}
    \mathcal{M}_c(x) = \mathbf{I}(x)
\end{equation}
\subsubsection{Rotation, Scaling and Opacity Map}
The remaining Gaussian parameters are related not only to the extracted features $\{\mathbf{f}^s\}_{s=1}^S$ in Sec.~\ref{sec:feature_extraction} but also to the spatial cues from estimated depth in Sec.~\ref{sec:depth_estimation}.
The former one provides a global context with image encoder $\mathcal{E}^{att}_{img}$ and the latter one should focus on structural details so that Gaussian parameters can be predicted in a feed-forward manner.
Hence, we construct an additional encoder $\mathcal{E}_{depth}$, which takes the depth map $\textbf{D}$ as input, to complement the coarse geometric awareness for each pixel.
The image features and the spatial features are fused by a U-Net like decoder $\mathcal{D}_{parm}$ to regress pixel-wise Gaussian features in full image resolution
\begin{equation}
\label{formula:decoder}
    \mathbf{\Gamma} = \mathcal{D}_{parm} (\mathcal{E}^{att}_{img}(\mathbf{I}) \oplus\mathcal{E}_{depth}(\mathbf{D}))
\end{equation}
where $\mathbf{\Gamma} \in \mathbb{R}^{H \times W \times D_G}$ is Gaussian features, $\oplus$ stands for concatenations at all feature levels.
The prediction heads, each composed of two convolution layers, are adapted to Gaussian features for specific Gaussian parameter map regression.
Before being used to formulate Gaussian representations, the rotation map should be normalized since it represents a quaternion
\begin{equation}
\label{formula:gs_rotation}
    \mathcal{M}_r(x) = Norm({h}_r(\mathbf{\Gamma}(x)))
\end{equation}
where ${h}_r$ is the rotation head.
The scaling map and the opacity map need activations to satisfy their range
\begin{align}
\begin{split}
\label{formula:gs_scaling&opacity}
    \mathcal{M}_s(x) = Softplus({h}_s(\mathbf{\Gamma}(x))) \\
    \mathcal{M}_\alpha(x) = Sigmoid({h}_\alpha(\mathbf{\Gamma}(x)))
\end{split}
\end{align}
where ${h}_s$ and ${h}_\alpha$ represent the scaling head and opacity head, respectively.
\subsubsection{Depth Residual Map}
\label{sec:position_res}
We discover that the estimated depth in Sec.~\ref{sec:depth_estimation} is still coarse, especially in the case of the absence of depth supervision during training.
To add high-frequency details onto coarse geometry, we further design a depth residual map 
\begin{equation}
\label{formula:gs_depth}
    \mathcal{M}_d(x) = \gamma Tanh({h}_d(\mathbf{\Gamma}(x)))
\end{equation}
where $h_d$ represents the depth residual head and we use the $Tanh$ function with a scaling factor $\gamma = 0.5$ to activate the predicted value in a small range.
Given the predicted depth map $\mathbf{D}$ in Eq.~\ref{formula:disp estimation} and the residual value in Eq.~\ref{formula:gs_depth}, a pixel located at $x$ can be immediately unprojected from image planes to 3D space using projection matrix $\mathbf{P} \in \mathbb{R}^{3 \times 4} $ structured with camera parameters $K$
\begin{equation}
\label{formula:gs_pisition}
    \mathcal{M}_p(x) = \Pi^{-1}_\mathbf{P}(x, \mathbf{D}(x)+\mathcal{M}_d(x))
\end{equation}
Finally, the learnable unprojection in Eq.~\ref{formula:gs_pisition} bridges 2D feature space and 3D Gaussian representation.

\subsection{Joint Training}
\label{sec:render}

The pixel-wise Gaussian parameter maps defined on both source views are then lifted to 3D space and aggregated to render photo-realistic novel view images using the Gaussian Splatting technique in Sec.~\ref{sec:preliminary}.
Our whole framework is fully differentiable so that we jointly train depth estimation (Sec.~\ref{sec:depth_estimation}) and Gaussian parameters prediction (Sec.~\ref{sec:gaussian_network}) which typically benefit each other.
The full pipeline can be trained with only rendering loss or a combination of depth loss and rendering loss when ground truth depth is available during training. 
When depth supervision is absent, we should also take into account the geometry consistency between two point clouds unprojected from left-view and right-view depth maps. 

\subsubsection{Training Loss}
\noindent\textbf{Rendering loss.} First, we use rendering loss composed of L1 loss and SSIM loss~\cite{wang2004ssim}, denoted as $\mathcal{L}_{mae}$ and $\mathcal{L}_{ssim}$ respectively, to measure the difference between the rendered and the ground truth image 
\begin{equation}
    \mathcal{L}_{render} = \lambda_1 \mathcal{L}_{mae} + \lambda_2 \mathcal{L}_{ssim}
\end{equation}
where we set $\lambda_1 = 0.8$ and $\lambda_2 = 0.2$ in our experiments. 

\noindent\textbf{Depth loss.} When ground truth depth is available, we minimize the L1 distance between the predicted and ground truth depth over the full sequence of predictions $\{\mathbf{d}^t\}_{t=1}^T$ with exponentially increasing weights, as shown in \cite{lipson2021raft-stereo}.
Given ground truth depth $\mathbf{d}_{gt}$, the loss is defined as
\begin{equation}
    \mathcal{L}_{depth} = \sum_{t=1}^{T} \mu^{T-t} \|\mathbf{d}_{gt} - \mathbf{d}^t\|_1
\end{equation}
where we set $\mu=0.9$ in our experiments.

\subsubsection{Geometry Regularization}
\label{sec:geo_regular}
Ground truth depth is not trivially accessible, especially for complex scene data. 
Only rendering loss can not ensure geometry consistency between the two input views.
Thus we try to minimize Chamfer distance between the unprojected Gaussian points of the two source views as a regularization term to boost the stereo-matching in two directions
\begin{align}
\begin{split}
    \mathcal{L}_{CD} = &\frac{1}{|\mathcal{P}^l|}\sum_{p_l \in \mathcal{P}^l} \min_{p_r \in \mathcal{P}^r} \|p_l - p_r\|_2 +\\
    &\frac{1}{|\mathcal{P}^r|}\sum_{p_r \in \mathcal{P}^r} \min_{p_l \in \mathcal{P}^l} \|p_r - p_l\|_2
\end{split}
\end{align}
where $\mathcal{P}^{l,r}$ represents a valid Gaussian point set from the left or right view.

Overall, the final loss function is defined as $\mathcal{L} = \mathcal{L}_{render} + \alpha \mathcal{L}_{CD} + \beta \mathcal{L}_{depth}$. In practice, we set $\beta = 0$ when ground truth depth is not available.

\newcommand{\tabledegree}{
\begin{table}
\caption{\textbf{Sensibility to camera sparsity on Human-only data.} We use the model trained under 8-camera setup to perform inference on a 6-camera setup.}
\small
\setlength\tabcolsep{2.0pt} 
\centering
\begin{tabular}{l|ccc|ccc}
\toprule
\multicolumn{1}{c}{\multirow{1}[6]{*}{Model}} & \multicolumn{3}{c}{8-camera setup} & \multicolumn{3}{c}{6-camera setup}\\

\cmidrule{2-7}   \multicolumn{1}{c}{} & PSNR$\uparrow$ & SSIM$\uparrow$ & \multicolumn{1}{c}{}{LPIPS$\downarrow$} & PSNR$\uparrow$ & SSIM$\uparrow$ & LPIPS$\downarrow$ \\ \midrule

FloRen~\cite{shao2022floren}     & 23.26 & 0.812 & 0.184 & 18.72 & 0.770 & 0.267\\
IBRNet~\cite{wang2021ibrnet} & 23.38 & 0.836 & 0.212 & 21.08 & 0.790 & 0.263\\
ENeRF~\cite{lin2022enerf} & 24.10 & 0.869 & 0.126 & 21.78 & 0.831 & 0.181\\
Ours & \textbf{25.57} & \textbf{0.898} & \textbf{0.112} & \textbf{23.03} & \textbf{0.884} & \textbf{0.168}\\

\bottomrule
\end{tabular}
\label{tab:camera}
\end{table}
}

\newcommand{\tableablation}{
\begin{table}[htpb!]
\small
\caption{\textbf{Quantitative ablation study on synthetic human-only data.} We report PSNR, SSIM and LPIPS metrics for evaluating the rendering quality, while the end-point-error (EPE) and the ratio of pixel error in 1 pix level for measuring depth accuracy. $\checkmark$ denotes training with depth supervision.}
\setlength\tabcolsep{3.0pt} 
\vspace{-2mm}
\centering
\begin{tabular}{l|c|ccc|cc}
\toprule
\multicolumn{1}{c}{\multirow{1}[5]{*}{Model}} & \multicolumn{1}{c}{\multirow{1}[5]{*}{\rotatebox{90}{\scriptsize{Dep.Sup.}\hspace{-0.8mm}}}} & \multicolumn{3}{c}{Rendering} & \multicolumn{2}{c}{Depth}\\

\cmidrule{3-7}   \multicolumn{1}{c}{} & \multicolumn{1}{c}{} & PSNR$\uparrow$ & SSIM$\uparrow$ & \multicolumn{1}{c}{}{LPIPS$\downarrow$} & EPE $\downarrow$ & 1 pix $\uparrow$ 
\\ 
\midrule

GPS-Gaussian     & $\checkmark$ & \textbf{25.05} & \textbf{0.886} & 0.121 & \textbf{1.494} & \textbf{65.94}\\
\textit{w/o} Joint Train. & $\checkmark$ & 23.97 & 0.862 & \textbf{0.115} & 1.587 & 63.71\\
\textit{w/o} Depth Enc. & $\checkmark$ & 23.84 & 0.858 & 0.204 & 1.496 & 65.87\\ \midrule

GPS-Gaussian    &  & 24.22 & 0.874 & 0.145 & 4.066 & 33.38 \\
GPS-Gaussian+   &  & 24.41 & 0.878 & 0.137 & 3.133 & 36.21 \\

\bottomrule
\end{tabular}
\label{tab:ablation}
\end{table}
}

\newcommand{\tableablationbg}{
\begin{table}[htpb!]
\small
\caption{\textbf{Quantitative ablation study on our captured human-scene data.} We report PSNR, SSIM and LPIPS metrics for evaluating the rendering quality.}
\setlength\tabcolsep{2.5pt} 
\centering
\vspace{-2mm}
\begin{tabular}{l|ccc}
\toprule

\multicolumn{1}{c}{\multirow{1}[0]{*}{Model}} & PSNR$\uparrow$ & SSIM$\uparrow$ & \multicolumn{1}{c}{}{LPIPS$\downarrow$} \\ \midrule

GPS-Gaussian               & 31.84 & 0.959 & 0.063 \\
\midrule
GPS-Gaussian+              & \textbf{33.74} & \textbf{0.971} & \textbf{0.041} \\
\textit{w/o} Geometry Reg. & 32.04 & 0.961 & 0.061 \\
\textit{w/o} Epipolar Att. & 32.00 & 0.960 & 0.059 \\
\textit{w/o} Depth Res. & 33.31 & 0.969 & 0.043 \\

\bottomrule
\end{tabular}
\vspace{-2mm}
\label{tab:ablation_bg}
\end{table}
}

\newcommand{\tableruntime}{
\begin{table}[t!]
\caption{\textbf{Run-time comparison.} We report the run-time correlated to the source views and each novel view on an RTX 3090 GPU. Input resolution is $512 \times 512$ for MVSplat, while all other methods take two $1024\times1024$ source images as input. Our methods can render multiple novel views concurrently in real-time.}
\small
\setlength\tabcolsep{1pt} 
\centering
\begin{tabular}{l|cc}
\toprule 
Methods & \makecell[c]{Source View\\Processing (ms)}\hspace{0.2cm}  & \makecell[c]{Novel View\\Rendering (ms/view)} \\ 
\midrule
FloRen~\cite{shao2022floren}  & 14  & 11    \\ 
IBRNet~\cite{wang2021ibrnet}  & 5   & 4000  \\
ENeRF~\cite{lin2022enerf}     & 11  & 125   \\
MVSplat~\cite{chen2024mvsplat}& 120 & 1.5   \\
GPS-Gaussian                  & 27  & 1.9   \\
GPS-Gaussian+                 & 30  & 1.9   \\
\bottomrule

\end{tabular}
\vspace{-2mm}
\label{tab:runtime}
\end{table}
}

\newcommand{\tablecomparebg}{
\begin{table*}[htpb!]
\small
\caption{\textbf{Quantitative comparison on human-scene datasets.} All methods are evaluated on an RTX 3090 GPU to report the speed of synthesizing one novel view with two $1024 \times 1024$ source images, except MVSplat~\cite{chen2024mvsplat} with two $512 \times 512$ images due to memory cost. Our method uses TensorRT for fast inference. $\dagger$ 4D-GS~\cite{Wu2024_4dgs} requires per-sequence optimization, while the other methods perform feed-forward inferences. The \colorbox{rred}{best}, the \colorbox{oorange}{second best} and the \colorbox{yyellow}{third best} are highlighted with different colors.}
\centering
\begin{tabular}{l|ccc|ccc|ccc|c}
\toprule
\multicolumn{1}{c}{\multirow{1}[6]{*}{Method}} & \multicolumn{3}{c}{DyNeRF~\cite{li2022neural}} & \multicolumn{3}{c}{ENeRF-outdoor~\cite{lin2022enerf}} & \multicolumn{3}{c}{Our Human-Scene Data} & \multirow{1}[6]{*}{FPS}\\

\cmidrule{2-10} \multicolumn{1}{c}{} & PSNR$\uparrow$ & SSIM$\uparrow$ & \multicolumn{1}{c}{LPIPS$\downarrow$} & PSNR$\uparrow$ & SSIM$\uparrow$  & \multicolumn{1}{c}{LPIPS$\downarrow$} & PSNR$\uparrow$ & SSIM$\uparrow$ & \multicolumn{1}{c}{LPIPS$\downarrow$} \\ \midrule

4D-GS~\cite{Wu2024_4dgs}$\dagger$  & 31.67 & \sbest 0.954 & \tbest 0.057 & 21.47 & 0.480 & 0.302 & 28.65 & 0.906 & 0.112 & /\\
MVSplat~\cite{chen2024mvsplat}     & 27.89 & 0.895 & 0.159 & 21.50 & \tbest 0.543 & 0.324 & 31.56 & 0.947 & 0.111 & \sbest 8 \\
IBRNet~\cite{wang2021ibrnet}       & \tbest 32.10 & 0.944 & 0.067 & \best 24.19 & \sbest 0.626 & \sbest 0.269 & \tbest 32.09 & \tbest 0.948 & \tbest 0.077 & 0.25 \\
ENeRF~\cite{lin2022enerf}          & \sbest 32.56 & \tbest 0.953 & \sbest 0.050 & \sbest 23.21 & 0.530 & \tbest 0.291 & \sbest 32.62 & \sbest 0.968 & \sbest 0.051 & \tbest 5 \\
GPS-Gaussian+                               & \best 33.72 & \best 0.961 & \best 0.039 & \tbest 23.07 & \best 0.643 & \best 0.238 & \best 33.74 & \best 0.971 & \best 0.041 & \best 25 \\
\bottomrule
\end{tabular}
\label{tab:num_compare_bg}
\end{table*}
}

\newcommand{\tablecompare}{
\begin{table*}[htpb!]
\small
\caption{\textbf{Quantitative comparison on human-only datasets.} All methods are evaluated on an RTX 3090 GPU to report the speed of synthesizing one novel view with two $1024 \times 1024$ source images. Our methods and FloRen~\cite{shao2022floren} use TensorRT for fast inference. $\dagger$ 3D-GS~\cite{kerbl2023_3dgs} requires per-subject optimization, while the other methods perform feed-forward inferences. The \colorbox{rred}{best}, the \colorbox{oorange}{second best} and the \colorbox{yyellow}{third best} are highlighted with different colors. $\checkmark$ denotes training with depth supervision.} 

\centering
\begin{tabular}{l|c|ccc|ccc|ccc|c}
\toprule
\multicolumn{1}{c}{\multirow{1}[6]{*}{Method}} & \multicolumn{1}{c}{\multirow{1}[6]{*}{\rotatebox{90}{\scriptsize{Dep.Sup.}\hspace{-0.8mm}}}} & \multicolumn{3}{c}{THuman2.0~\cite{yu2021function4d}} & \multicolumn{3}{c}{Twindom~\cite{twindom}} & \multicolumn{3}{c}{Human-Only Real Data} & \multirow{1}[6]{*}{FPS}\\

\cmidrule{3-11} \multicolumn{1}{c}{} & & PSNR$\uparrow$ & SSIM$\uparrow$ & \multicolumn{1}{c}{LPIPS$\downarrow$} & PSNR$\uparrow$ & SSIM$\uparrow$  & \multicolumn{1}{c}{LPIPS$\downarrow$} & PSNR$\uparrow$ & SSIM$\uparrow$ & \multicolumn{1}{c}{LPIPS$\downarrow$} \\ \midrule

3D-GS~\cite{kerbl2023_3dgs}$\dagger$   &  & \tbest 24.18 & 0.821 & 0.144 & 22.77 & 0.785 & 0.153 & 22.97 & 0.839 & 0.125 & / \\
FloRen~\cite{shao2022floren}  & \checkmark & 23.26 & 0.812 & 0.184 & 22.96 & 0.838 & 0.165 & 22.80 & 0.872 & 0.136 & \tbest 15\\
IBRNet~\cite{wang2021ibrnet}  &            & 23.38 & 0.836 & 0.212 & 22.92 & 0.803 & 0.238 & 22.63 & 0.852 & 0.177 & 0.25\\
ENeRF~\cite{lin2022enerf}     &            & 24.10 & \tbest 0.869 & \sbest 0.126 & \tbest 23.64 & \tbest 0.847 & \sbest 0.134 & \tbest 23.26 & \tbest 0.893 & \tbest 0.118 & 5\\
\midrule
GPS-Gaussian    & \checkmark & \best 25.57 & \best 0.898 & \best 0.112 & \best 24.79 & \best 0.880 & \best 0.125 & \best 24.64 & \best 0.917 & \best 0.088 & \best 25\\
GPS-Gaussian+   &  & \sbest 24.72 & \sbest 0.894 & \tbest 0.129 & \sbest 24.23 & \sbest 0.871 & \tbest 0.141 & \sbest 23.45 & \sbest 0.904 & \sbest 0.106 & \sbest 24 \\
\bottomrule
\end{tabular}
\label{tab:num_compare}
\end{table*}
}

\tablecomparebg
\begin{figure*}
  \centering
  \vspace{-2mm}
  \includegraphics[width=\textwidth]{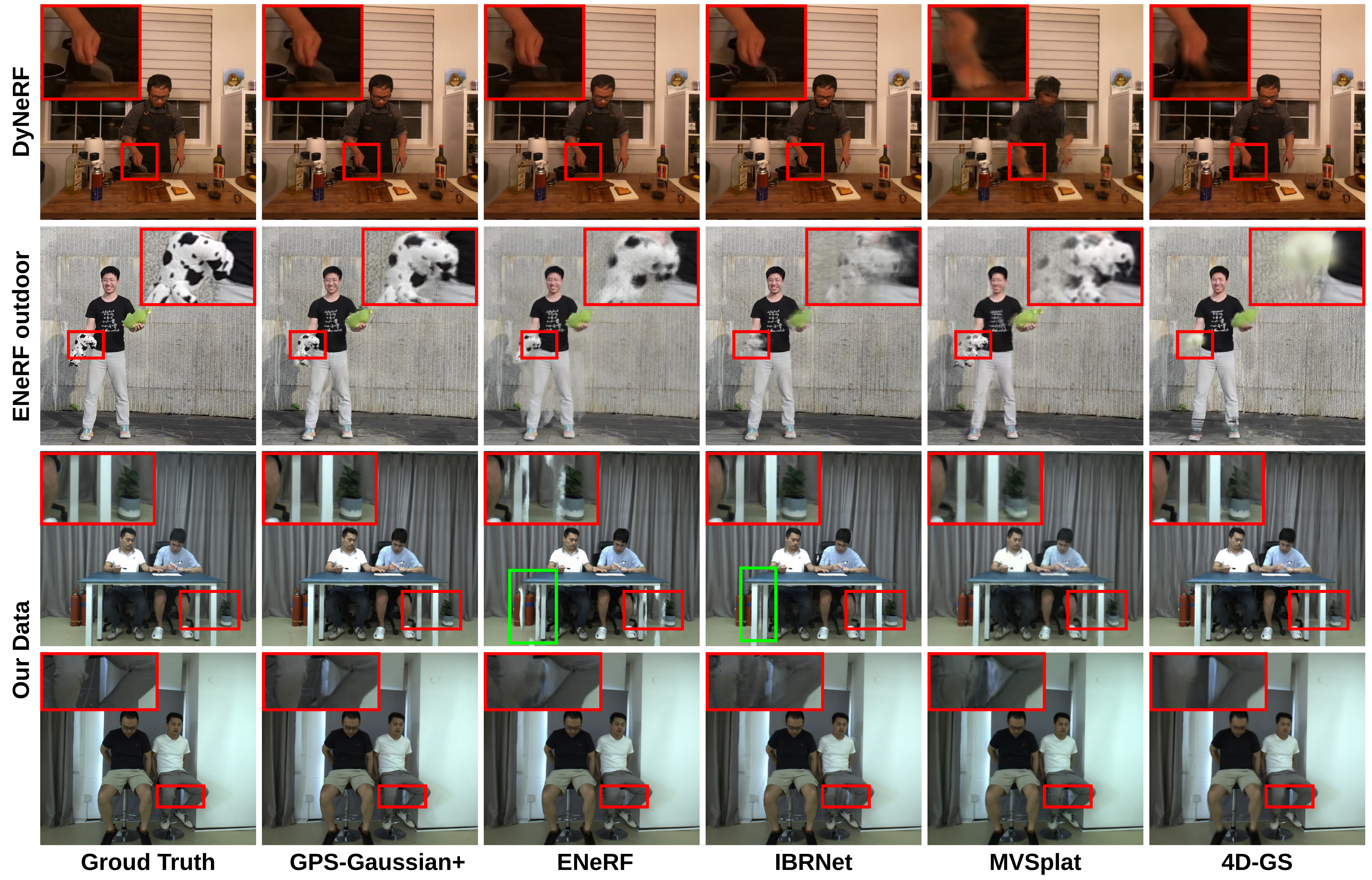}
  \vspace{-5mm}
  \caption{\textbf{Qualitative comparison on human-scene data.} Our method produces high-quality renderings with respect to others.}
  \label{fig:comparison_bg}
  \vspace{-3mm}
\end{figure*}

\section{Experiments}
\label{sec:experiments}

\subsection{Datasets and Metrics} 
\label{sec:dataset}
\noindent\textbf{Human-Scene data.}
To train and evaluate our network, we collect character performance data in the scene from DyNeRF~\cite{li2022neural} and ENeRF-outdoor~\cite{lin2022enerf} datasets. We take 4 motion sequences from DyNeRF, each of which contains 300 frames. The first 220 frames are used as training data, and we evaluate our method on the rest of the data.
For ENeRF-outdoor data, we take 4 motion sequences of 300 frames as training data, and 2 motion sequences of unseen characters as test data.
We also capture motion sequences of single-character or multiple-character performance in 3 different scenes with a forward-facing camera rig, as shown in Fig.~\ref{fig:camera}(a), to test the robustness of our method across scenes.
In particular, 10 cameras are positioned in a line, spanning 1.6 meters.
Four cameras with red circles in Fig.~\ref{fig:camera}(a), are used as inputs which compose 3 pairs of source views, and the others serve as novel views during validation.
For rendering continuity across source-view pairs, we use all 10 views as supervision during training.
For each scene, our captured dataset consists of 3 sequences for training and 2 sequences for test, so there are 15 sequences in total.
We train a model on each dataset and the models can generalize to unseen characters in the scene. 
For our captured data, our model is able to handle all three backgrounds.
In terms of human-scene ratio, our data and ENeRF-outdoor capture full-body characters with small-focal cameras, while DyNeRF focuses on upper-body.
Due to the original resolution of raw data, we set all images to $1K$ resolution.

\noindent\textbf{Human-only data.}
To learn human priors from a large amount of data, we collect 1700 and 526 human scans from Twindom~\cite{twindom} and THuman2.0~\cite{yu2021function4d}, respectively.
We randomly select 200 and 100 scans as validation data from Twindom and THuman2.0, respectively.
In addition, we uniformly position 8 cameras in a cycle, thus the angle between two neighboring cameras is about $45^{\circ}$. 
To test the robustness in real-world scenarios, we capture real data of 4 characters in the same 8-camera setup and prepare 8 additional camera views for evaluation, as shown in Fig.~\ref{fig:camera}(b). 
For synthetic data, we render images on $2K$ resolution as rendering supervision during training and as ground truth during the test.

\noindent\textbf{Evaluation metric.}
Following ENeRF~\cite{lin2022enerf}, we evaluate our method and other baselines with PSNR, SSIM~\cite{wang2004ssim} and LPIPS~\cite{zhang2018lpips} as metrics for the rendering results in valid regions of novel views. 
\subsection{Implementation Details} 
Our method is trained on a single RTX3090 graphics card using AdamW~\cite{adamw} optimizer with an initial learning rate of $2e^{-4}$.
For real-captured human-scene data, we train the whole network from scratch for around $100k$ iterations with rendering loss and Chamfer distance.
For DyNeRF~\cite{li2022neural} data, we set $\alpha = 0.005$ for its large range of depth, while we set $\alpha = 0.5$ for ENeRF-outdoor~\cite{lin2022enerf} and our captured data.
For synthetic human-only data, THuman2.0~\cite{yu2021function4d} and Twindom~\cite{twindom}, we have 2 strategies to train networks. 
We can still train the whole network from scratch for $100k$ iterations with rendering loss and Chamfer distance. 
When depth information is available during training, the depth estimation module can be firstly trained for $40k$ iterations and we then jointly train depth estimation and Gaussian parameters prediction for $100k$ iterations. 

\subsection{Comparisons}
\noindent\textbf{Baselines.}
Considering that our goal is instant novel view synthesis, we compare our GPS-Gaussian+ against generalizable methods including Gaussian Splatting-based method MVSplat~\cite{chen2024mvsplat}, implicit method ENeRF~\cite{lin2022enerf}, image-based rendering method FloRen~\cite{shao2022floren} and hybrid method IBRNet~\cite{wang2021ibrnet}. 
In particular, it is difficult to train MVSplat on masked human-only data for its probabilistic modeling and FloRen relies on human prior, so we compare MVSplat only on human-scene data and FloRen only on human-only data. 
Following our setting, all baselines are trained on the same training set from scratch and take two source views as input for synthesizing the targeted novel view.
The preliminary work, GPS-Gaussian, and FloRen use ground truth depths of synthetic human-only data for supervision.
We further prepare the comparison with the original 3D-GS~\cite{kerbl2023_3dgs} for static human-only data and with 4D-GS~\cite{Wu2024_4dgs} for sequential human-scene data which are optimized on all input views using the default strategies in the released code.

\tablecompare
\begin{figure*}[h!]
  \centering
  \includegraphics[width=0.94\textwidth]{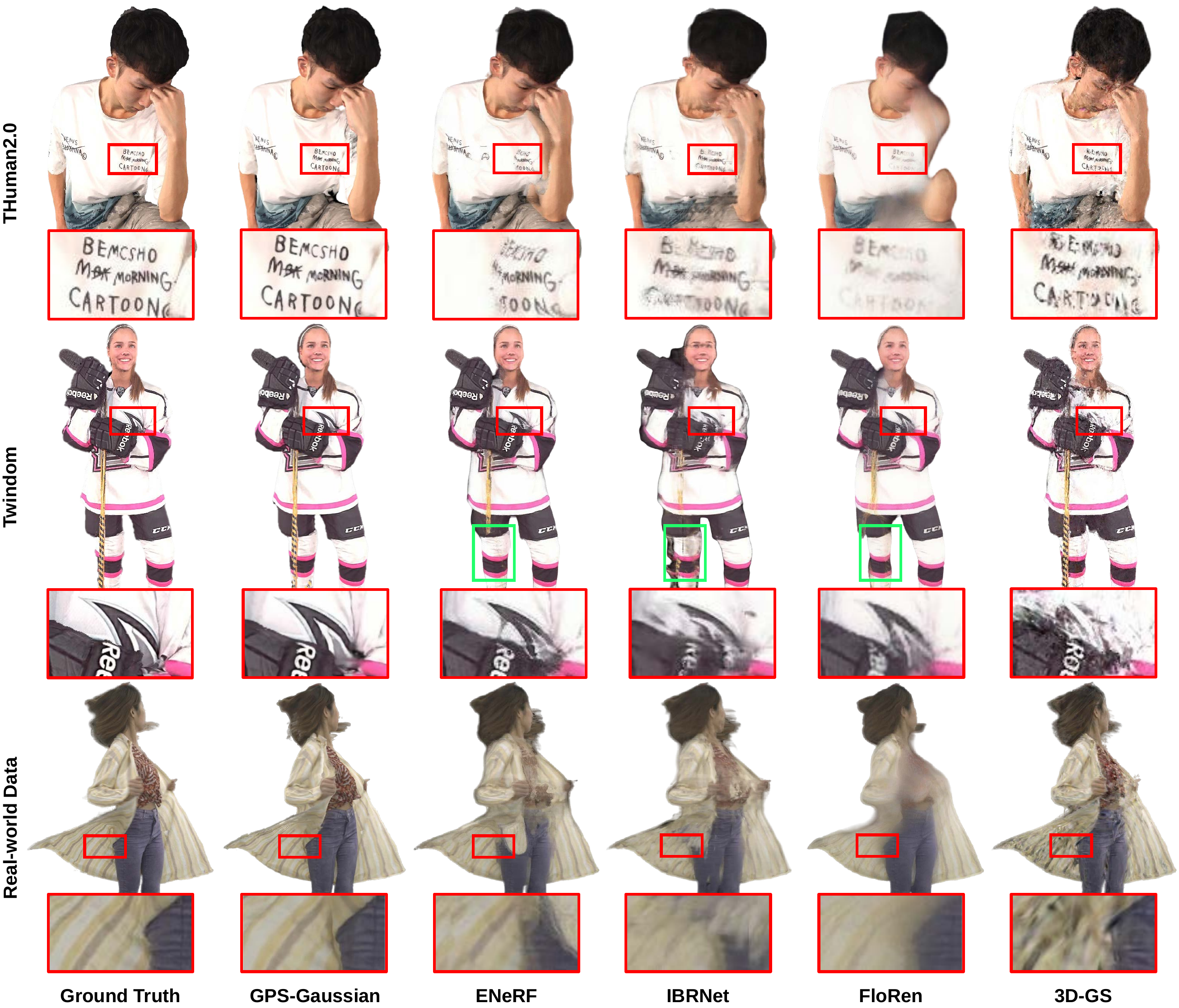}
  \caption{\textbf{Qualitative comparison on human-only data.} Our method produces more detailed human appearances and can recover more reasonable geometry.}
  \label{fig:comparison}
  \vspace{-3mm}
\end{figure*}

\subsubsection{Results on Human-Scene Data}
We compare state-of-the-art methods on 3 real captured human-scene datasets. 
In Table~\ref{tab:num_compare_bg}, our approach achieves superior or competitive results at the fastest speed with respect to other methods.
In particular, our approach makes a great improvement on metric LPIPS which reveals better global rendering quality. 
We notice that camera parameters are not perfectly calibrated in the ENeRF-outdoor~\cite{lin2022enerf} dataset.
Although the bad calibration has a tough impact on the back-propagation of rendering loss, our method can still synthesize fine-grained novel view images with more detailed appearances in Fig.~\ref{fig:comparison_bg}. 
Due to the lack of consistent geometry prior, ENeRF and IBRNet can easily make floating artifacts illustrated in Fig.~\ref{fig:comparison_bg}.
MVSplat and ENeRF rely on multiplane sweeping to infer geometry from sparse views, such representation can hardly handle thin structure object, \textit{e.g.} knife in Fig.~\ref{fig:comparison_bg}. 
Although 4D-GS accelerates the optimization process from the original 3D-GS by decomposing spatial-temporal deformation into multi-resolution planes, such decomposition produces blurry results under fast movements, see Fig.~\ref{fig:teaser} and Fig.~\ref{fig:comparison_bg}.
Thanks to determinant stereo-matching and geometry regularization, our generated geometry is consistent from the two source views so that our rendering results are more decent with fewer floating artifacts.

\tableablation
\tableablationbg

\subsubsection{Results on Human-Only Data}
We illustrate comparisons on two synthetic datasets and our collected real-world data in Table~\ref{tab:num_compare}.
Our method outperforms all baselines on all metrics and achieves a much faster rendering speed. 
Once occlusion occurs, some target regions under the novel view are invisible in one or both of the source views.
ENeRF and IBRNet are not able to render reasonable results due to depth ambiguity. 
The unreliable geometric proxy in these cases also makes FloRen produce blurry outputs even if it employs the depth and flow refining networks.
In our method, the efficient stereo-matching strategy and the geometry regularization help to alleviate the adverse effects caused by occlusion.
In addition, it takes several minutes for 3D-GS parameter optimization for a single frame and produces noisy renderings of novel views, see Fig.~\ref{fig:comparison}, from such sparse views.
Also, we demonstrate the effectiveness of our method on thin structures, such as the hockey stick and robe in Fig.~\ref{fig:comparison}.

\subsection{Ablation Studies}
\label{sec:ablation}

In this part, we evaluate the effectiveness of our proposed components in GPS-Gaussian and GPS-Gaussian+ through ablation studies.
The efficacy of joint training and depth encoder proposed in GPS-Gaussian are validated on aforementioned synthetic human-only data. 
As depth information is accessible in synthetic data, we further evaluate depth (identical to disparity) estimation, other than rendering metrics, with the end-point-error (EPE) and the ratio of pixel error in 1 pix level, following \cite{lipson2021raft-stereo}. 
Furthermore, the additional components in GPS-Gaussian+ are evaluated on our captured human-scene data because our data includes single and multiple characters in different scenes.
We also conduct a comparison between GPS-Gaussian and GPS-Gaussian+ on both datasets in order to illustrate the efficiency of adaptive integration under the setting of training without depth supervision.

\begin{figure}[!t]
  \centering
  \includegraphics[width=\linewidth]{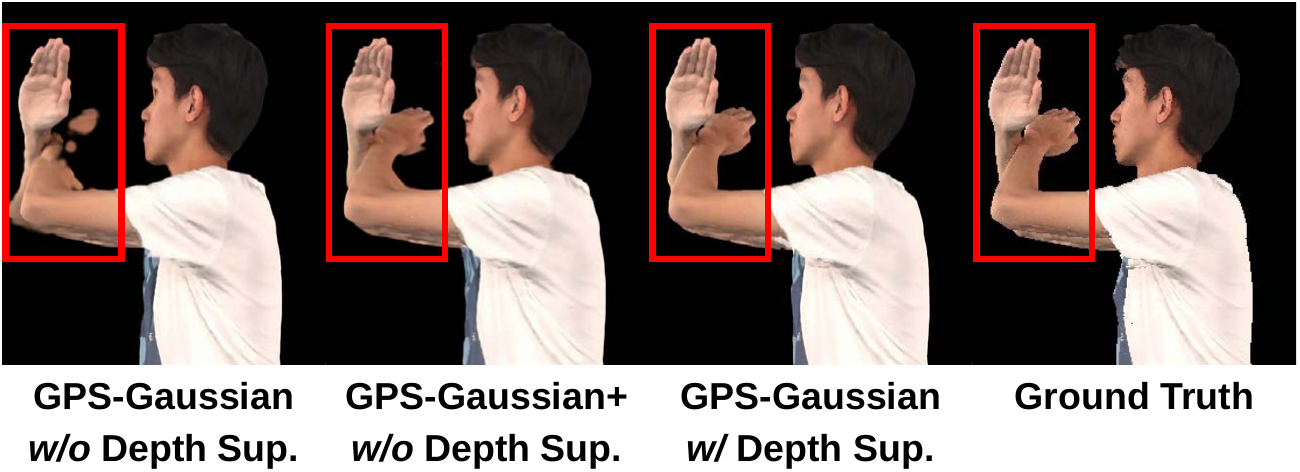}
  \caption{\textbf{Qualitative ablation study on GPS-Gaussian/GPS-Gaussian+ with different supervision settings.} We show the effectiveness of the integration in GPS-Gaussian+ when neglecting depth supervision.}
  \label{fig:ablation2}
  \vspace{-4mm}
\end{figure}

\begin{figure}[!t]
  \centering
  \includegraphics[width=\linewidth]{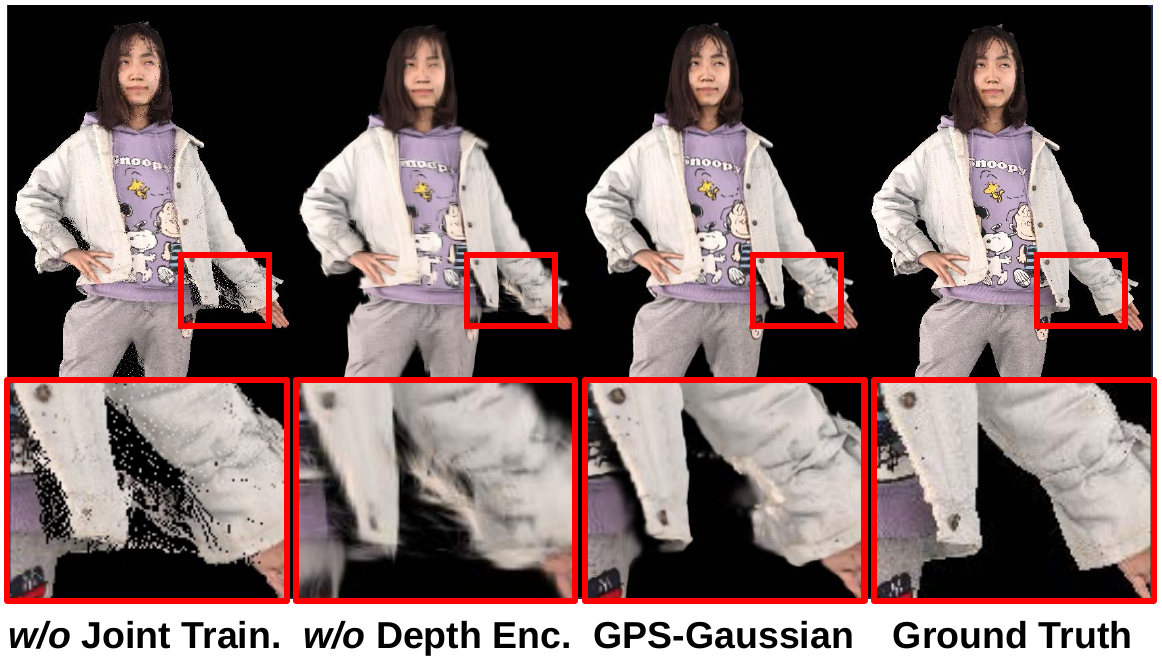}
  \caption{\textbf{Qualitative ablation study on designed components of GPS-Gaussian.} We show the effectiveness of the joint training and the depth encoder in the full pipeline. 
  The proposed designs make the rendering results more visually appealing with fewer artifacts and less blurry.}
  \label{fig:ablation}
  \vspace{-4mm}
\end{figure}

\begin{figure*}[htpb!]
  \centering
  \includegraphics[width=0.95\linewidth]{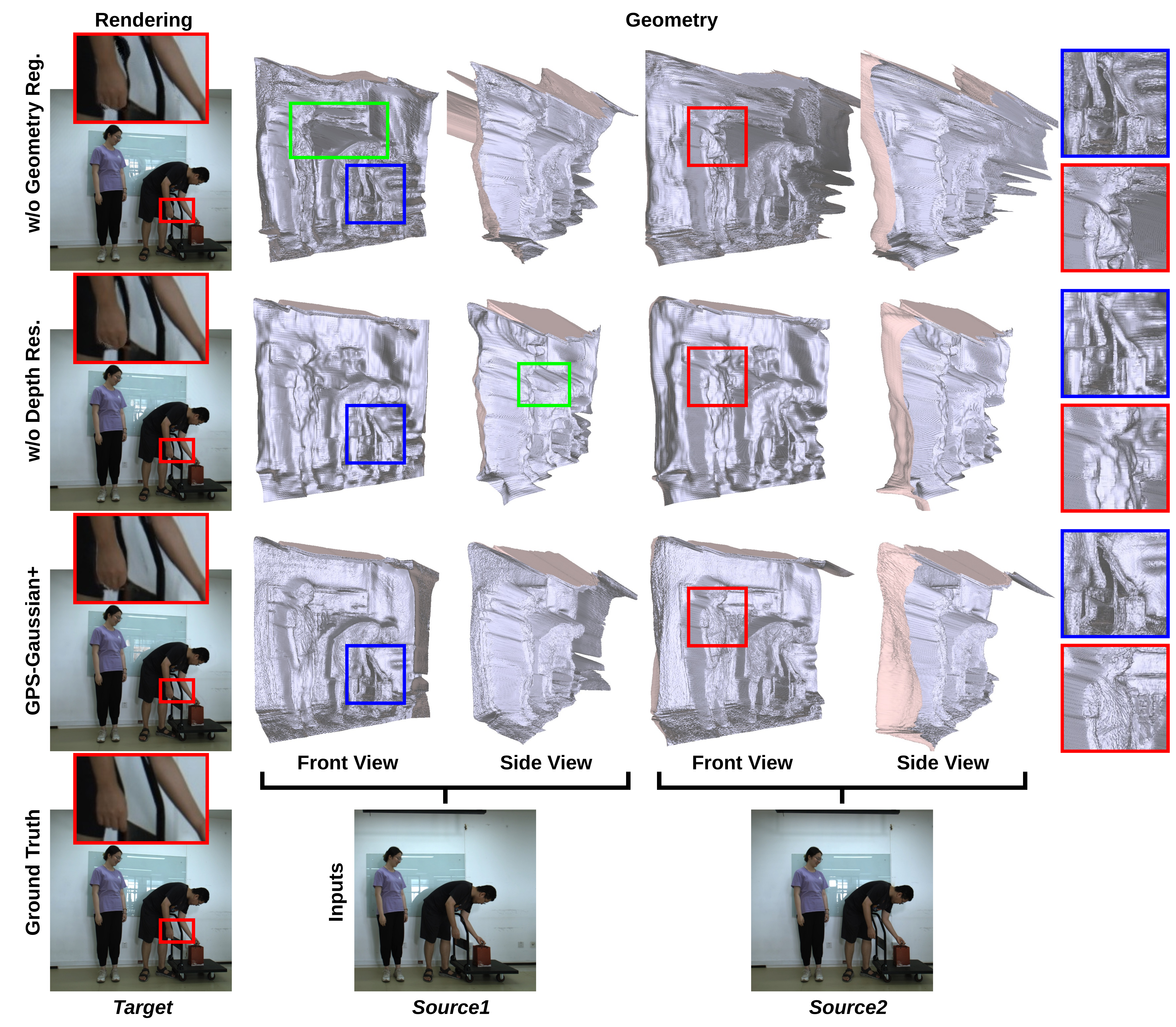}
  \vspace{-2mm}
  \caption{\textbf{Qualitative ablation study on designed components in GPS-Gaussian+ for geometry.} We show the effectiveness of the geometry regularization and the depth residual in the full pipeline for geometry reconstruction.}
  \vspace{-2mm}
  \label{fig:ablation_geo}
\end{figure*}

\subsubsection{Effects of Depth Supervision}
A consistent geometry prior is essential to rendering reasonable images in novel views, especially for explicit Gaussian Splatting.
Although the rendering metrics in Table~\ref{tab:ablation} are not dramatically degraded for GPS-Gaussian when neglecting depth supervision, the geometry metrics are not competitive with respect to the models using depth supervision.
In addition, a wrong geometry prior could produce unreasonable rendering results in occluded regions, see Fig.~\ref{fig:ablation2}.
Therefore, how to exploit accurate geometry proxy in the case of the lack of depth supervision is the key to our extensions. 

\subsubsection{Effects of Joint Training Mechanism}
For GPS-Gaussian trained on synthetic human-only data, we compare jointly training both depth estimation and rendering modules with separately training them.
We design a model substituting the differentiable Gaussian rendering with point cloud rendering at a fixed radius.
Since the point cloud renderer is no longer differentiable, the rendering quality is merely based on the accuracy of depth estimation while the rendering loss could not conversely promote the depth estimator.
The rendering results in Fig.~\ref{fig:ablation} witness floating artifacts due to the depth ambiguity in the margin area of the source views where the depth value changes drastically.
In Table~\ref{tab:ablation}, joint training makes a more robust depth estimator with a $5\%$ improvement in EPE.

\subsubsection{Effects of Depth Encoder}
We claim that merely using image features is insufficient for predicting Gaussian parameters.
Herein, we ablate the depth encoder from our model, thus the Gaussian parameter decoder only takes as input the image features to predict $\mathcal{M}_r, \mathcal{M}_s, \mathcal{M}_\alpha$ simultaneously.
As shown in Fig.~\ref{fig:ablation}, the ablated model fails to recover the details of human appearance, leading to blurry rendering results.
The scale of Gaussian points is impacted by comprehensive factors including depth, texture and surface roughness, see Sec.~\ref{sec:vis_scaling}.
The absence of spatial awareness degrades the regression of scaling map $\mathcal{M}_s$, which deteriorates the visual perception reflected on LPIPS, even with a comparable depth estimation accuracy, as shown in Table~\ref{tab:ablation}.

\subsubsection{Effects of Geometry Regularization}
\begin{figure}[h!]
  \centering
  \includegraphics[width=\linewidth]{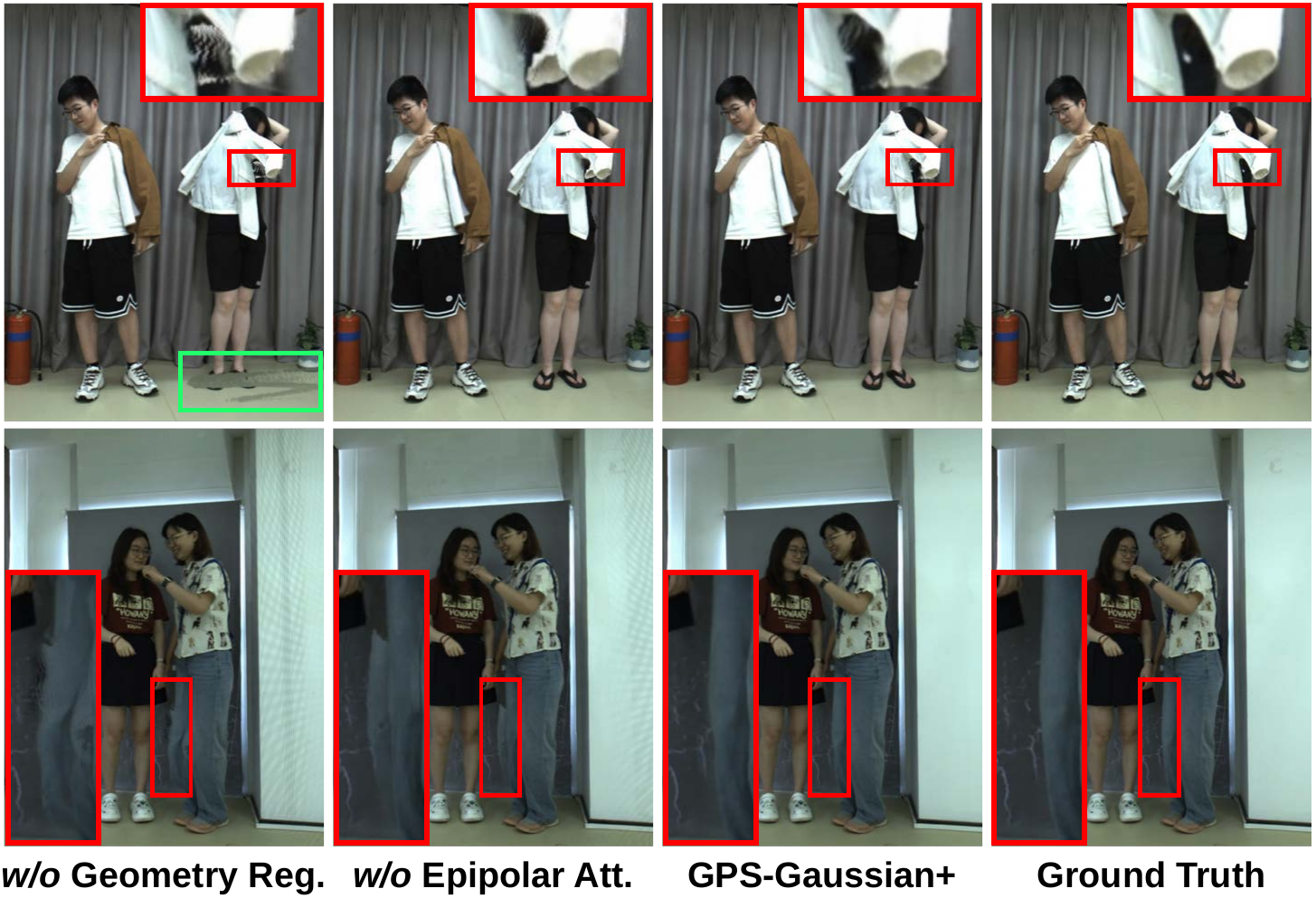}
  \caption{\textbf{Qualitative ablation study on designed components in GPS-Gaussian+ for rendering.} We show the effectiveness of the geometry regularization and the epipolar attention in the full pipeline.}
  \label{fig:ablation_bg}
  \vspace{-2mm}
\end{figure}

In GPS-Gaussian+ trained on real captured data without depth supervision, geometry regularization is designed to preserve geometry consistency between Gaussian points of the two source views.
Due to the lack of depth supervision, marginal regions in Fig.~\ref{fig:ablation_bg} and regions with view-dependent reflection in Fig.~\ref{fig:ablation_geo} are hardly reconstructed when missing our proposed geometry regularization. 
Such geometric constraints can boost the unsupervised depth learning of the two source views to reach a mutual optimum.
A good geometry prior also improves the rendering results, as reported in Table~\ref{tab:ablation_bg}.

\subsubsection{Effects of Depth Residual}
Without depth supervision, GPS-Gaussian generates scarcely reasonable geometry, as shown in the fourth row of Table~\ref{tab:ablation}.
Although GPS-Gaussian+ integrates the aforementioned adaptions, the geometry of the second row in Fig.~\ref{fig:ablation_geo} is still not acceptable.
This problem is caused by two reasons. First, a downsampling operator is used in the stereo-matching module to minimize time cost. Second, the differentiability of point position in Gaussian Splatting is not satisfactory enough. 
By using our proposed depth residual map, our model recovers more details and corrects geometry artifacts, see the third row of Fig.~\ref{fig:ablation_geo}.

\subsubsection{Effects of Epipolar Attention}
Compared with GPS-Gaussian, GPS-Gaussian+ incorporates epipolar attention into the feature extraction module to achieve a solid stereo-matching result with only rendering loss.
Epipolar attention allows the encoder to exchange useful information between source views so that we can build a compact cost volume for stereo-matching.
Even if the disparity (identical to depth) is not accessible during training, our proposed model corrects floating artifacts caused by wrong matching in Fig.~\ref{fig:ablation_bg}. 
Since we apply such attention mechanism only along epipolar line, the time cost is on par with GPS-Gaussian, see Table~\ref{tab:runtime}.

In general, our adaptive integration is designed to compensate for the absence of depth supervision.
In Table~\ref{tab:ablation}, GPS-Gaussian+ without depth supervision achieves competitive rendering results against GPS-Gaussian with depth supervision and produces reasonable geometry results.
Moreover, the adaptive integration works better on full-scene images than masked human-only images. 
When the background is concerned, GPS-Gaussian+ can largely improve all rendering metrics with respect to GPS-Gaussian, as shown in Table~\ref{tab:ablation_bg}.

\subsection{Visualization of Opacity Maps}
\label{sec:vis_opacity}
\begin{figure}[!t]
  \centering
  \includegraphics[width=0.5\textwidth]{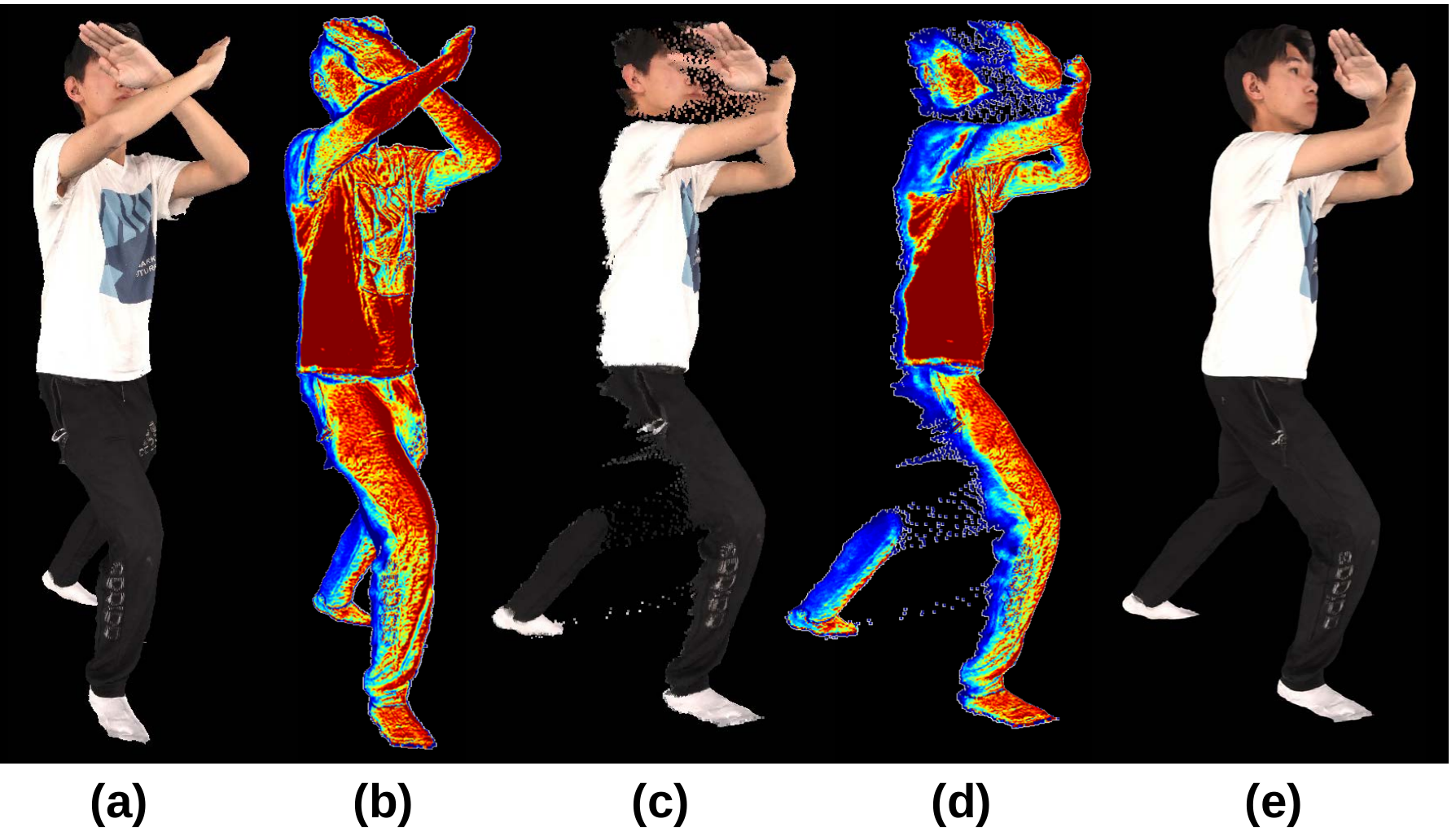}
  \vspace{-4mm}
  \caption{\textbf{Visualization of opacity maps.} (a) One of the source view images. (b) The predicted opacity map related to (a). (c)/(d) The directly projected color/opacity map at novel viewpoint. (e) Novel view rendering results. A cold color in (b) and (d) represents an opacity value near 0, while a hot color near 1. The low opacity values predicted for the outliers make them invisible.}
  \label{fig:vis_opacity}
  \vspace{-2mm}
\end{figure}

We discover that the joint regression with Gaussian parameters eliminates the outliers by predicting an extremely low opacity for the Gaussian points centered at these positions.
The visualization of opacity maps is shown in Fig.~\ref{fig:vis_opacity}.
Since the depth prediction works on low resolution and upsampled to full image resolution, the drastically changed depth in the margin areas causes ambiguous predictions (\emph{e.g.} the front and rear placed legs and the crossed arms in Fig.~\ref{fig:vis_opacity}).
These ambiguities lead to rendering noise on novel views when using a point cloud rendering technique.
Thanks to the learned opacity map, the low opacity values make the outliers invisible in novel view rendering results, as shown in Fig.~\ref{fig:vis_opacity}~(e).

\begin{figure}[!t]
  \centering
  \includegraphics[width=\linewidth]{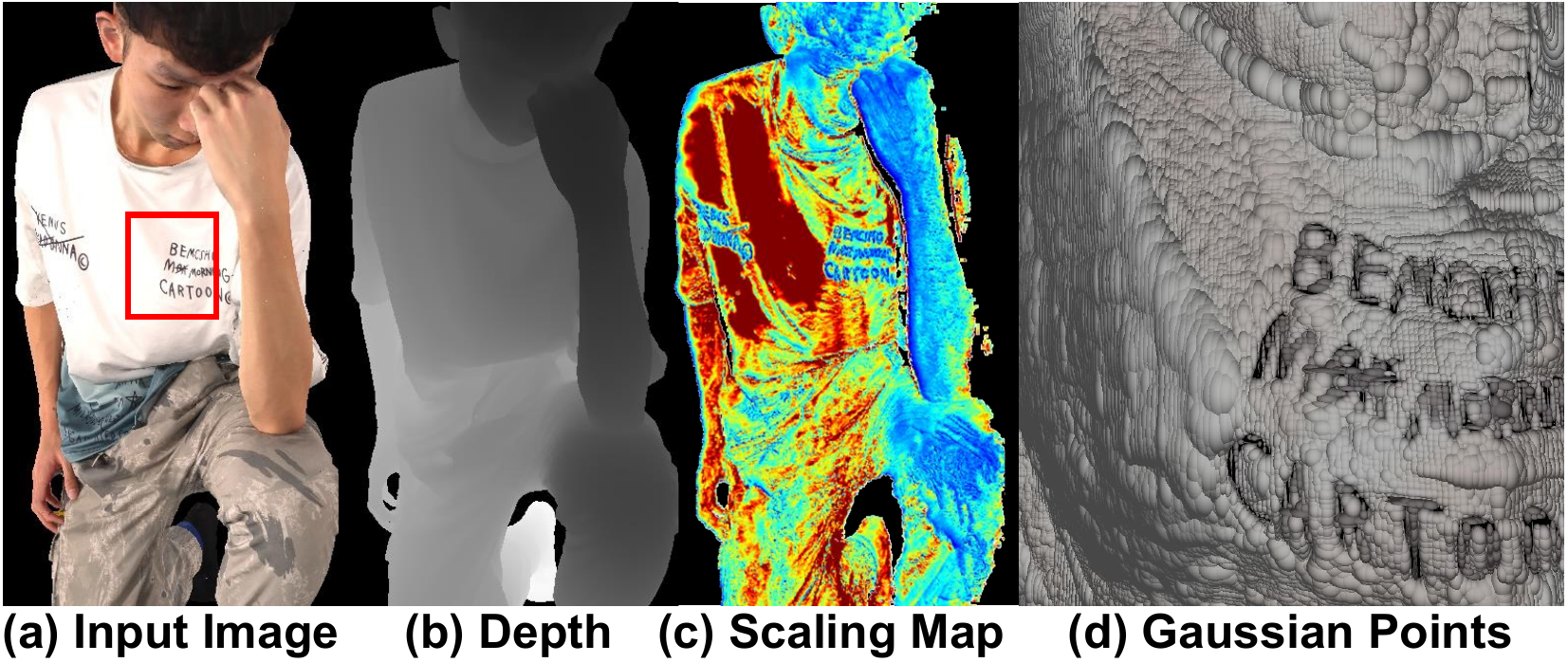}
  \vspace{-4mm}
  \caption{\textbf{Visualization of scaling map and the shape of Gaussian points.} (a) One of the source view images. (b) The depth of (a). (c) The scaling map shown in heat map, where a hotter color represents a larger value. (d) The zoom-in Gaussian points of the boxed area in (a). The depth and scaling map are normalized.}
  \label{fig:vis_scaling}
  \vspace{-2mm}
\end{figure}

\subsection{Visualization of Scaling Maps}
\label{sec:vis_scaling}
The visualization of the scaling map (mean of three axes) in Fig.~\ref{fig:vis_scaling}~(c) indicates that the Gaussian points with lower depth roughly have smaller scales than the distant ones.
However, the scaling property is also impacted by comprehensive factors.
For example, as shown in Fig.~\ref{fig:vis_scaling}~(c) and (d), fine-grained textures or high-frequency geometries lead to small-scaled Gaussians.

\begin{figure}
  \centering
  \includegraphics[width=\linewidth]{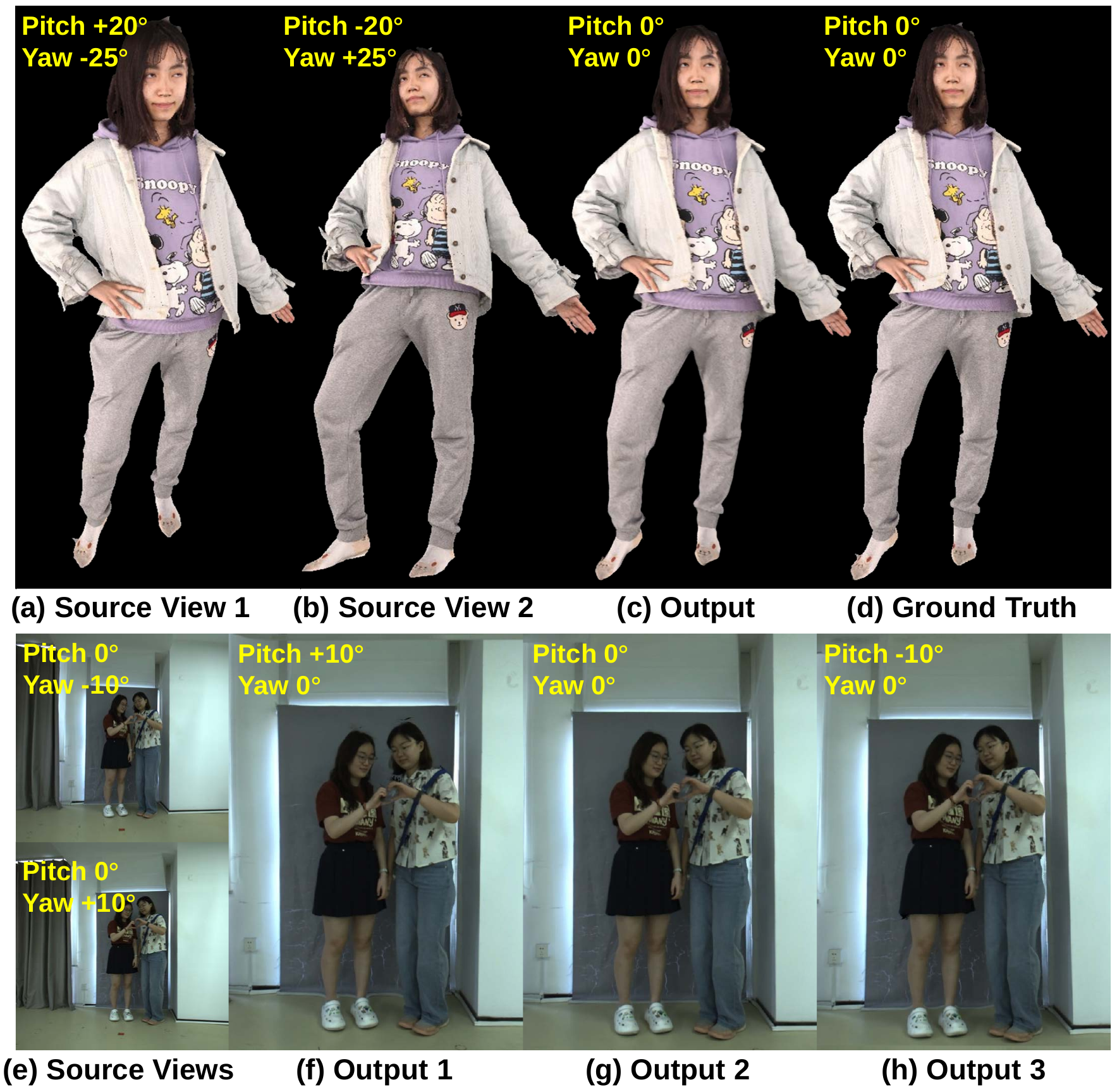}
  \vspace{-4mm}
  \caption{\textbf{Results on random camera views.} In the first row, (a) and (b) are the source view images with an extreme pitch and yaw, (c) is the novel view rendering result and (d) is the novel view ground truth. In the second row, we show (e) source views and target-view renderings with pitch angles of (f) $+10^\circ$, (g) $0^\circ$ and (h) $-10^\circ$.}
  \label{fig:pitching}
  \vspace{-2mm}
\end{figure}

\subsection{Robustness to Random Camera Views}
\label{sec:random_setup}
We evaluate the robustness of our method to the randomly placed source-view cameras in the first row of Fig.~\ref{fig:pitching}.
The model trained under a uniformly placed 8-camera setup in Sec.~\ref{sec:experiments} shows a strong generalization capability to random camera setup with a pitch in range of [$-20^\circ, +20^\circ$] and yaw in range of [$-25^\circ, +25^\circ$] for human-only data.
In Fig.~\ref{fig:pitching} (f) and (h), our method achieves reasonable renderings of novel views with a pitch angle of about $\pm 10^\circ$ for human-scene data, even without any supervision of views with pitch angles during training.

\subsection{Robustness to Unseen Scenes}
\label{sec:unseen_scene}
\begin{figure}[t!]
  \centering
  \includegraphics[width=\linewidth]{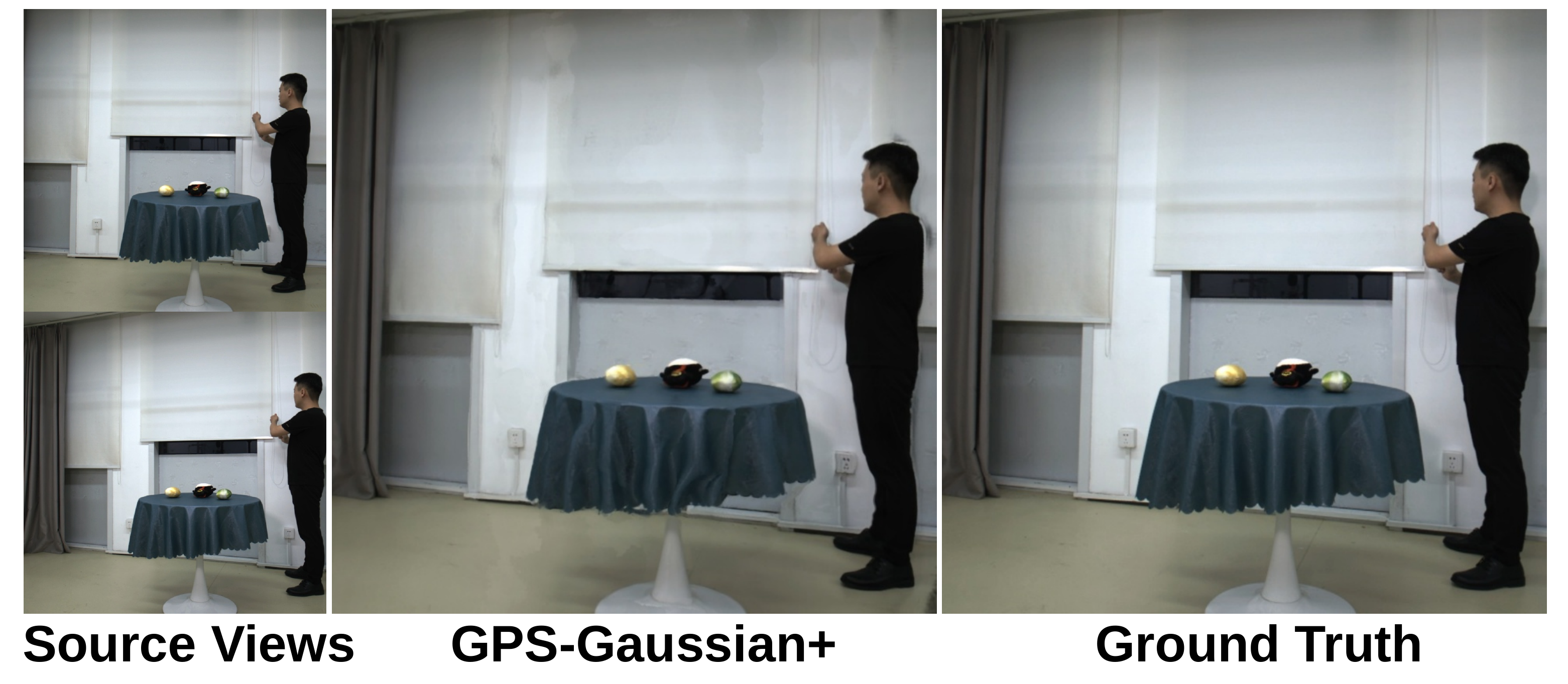}
  \vspace{-4mm}
  \caption{\textbf{Result on unseen scenes.} In the case of unseen background during training, our method achieves reasonable rendering without any finetuning.}
  \label{fig:unseen_scene}
  \vspace{-4mm}
\end{figure}

We further evaluate the generalization ability of our approach on unseen scene data in Fig.~\ref{fig:unseen_scene}.
We use the aforementioned model (Sec.~\ref{sec:dataset}) trained on our captured data under three backgrounds without any fine-tuning.
Even if the background in Fig.~\ref{fig:unseen_scene} is totally unseen during training, our method is able to generate reasonable renderings.

\subsection{Comparison on Run-Time}
\label{sec:run-time}

We conduct all experiments of our method and other baseline methods on the same machine with an RTX 3090 GPU of 24GB memory, except memory-consuming MVSplat~\cite{chen2024mvsplat}.
Even if we prepare another machine with a V100 GPU of 32GB memory, MVSplat can only be fed with input images of $512 \times 512$ resolution during training.
In Table~\ref{tab:runtime}, the overall run-time can be generally divided into two parts: one correlating to the source views and the other concerning the desired novel view.
The source view correlated computation in FloRen~\cite{shao2022floren} refers to coarse geometry initialization while the key components, the depth and flow refinement networks, operate on novel viewpoints.
IBRNet~\cite{wang2021ibrnet} uses transformers to aggregate multi-view cues at each sampling point aggregated to the novel view image plane, which is time-consuming.
ENeRF~\cite{lin2022enerf} constructs two cascade cost volumes on the target viewpoint, then predicts the target view depth followed by a depth-guided sampling for volume rendering.
Once the target viewpoint changes, these methods need to recompute the novel view correlated modules.
However, the computation on source views dominates the run-time of GPS-Gaussian+, which includes binocular depth estimation and Gaussian parameter map regression.
Similar to our method, MVSplat~\cite{chen2024mvsplat} spends the majority of run-time on the multi-view stereo process of source views, which includes multiple CNN and transformer operations. 
As reported in Table~\ref{tab:runtime}, it takes only $1.9$ ms to render the 3D Gaussians to the desired novel view of human-scene data for GPS-Gaussian+, while this can be reduced to $0.8$ ms when rendering human-only data with fewer Gaussian points.
This allows us to render multiple novel views simultaneously, which caters to a wider range of applications such as holographic displays.
Suppose that $n=10$ novel views are required concurrently, it takes our method $T = T_{src} + n \times T_{novel} = 49ms$ to synthesize, while $135ms$ for MVSplat and $1261ms$ for ENeRF.
In a real-world capture system, we should also consider I/O process and human matting, thus the frame rate is slightly degraded in Table~\ref{tab:num_compare_bg} and Table~\ref{tab:num_compare}.

\tableruntime

\begin{figure}[t]
  \centering
  \includegraphics[width=\linewidth]{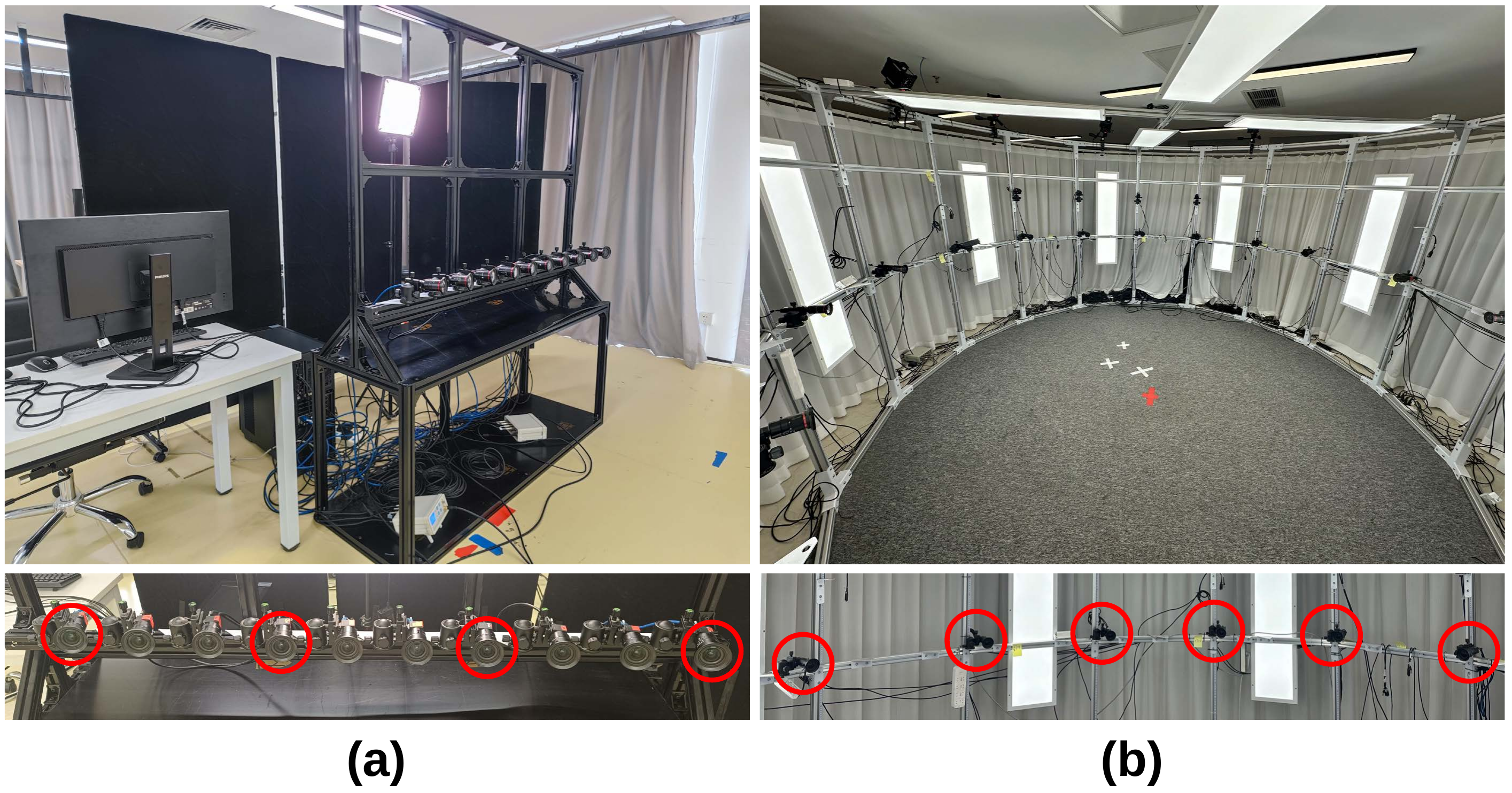}
  \caption{\textbf{Capture systems.} We show (a) forward-facing camera rig capturing human-scene data and (b) human-centered camera stage for human-only data.}
  \label{fig:camera}
\end{figure}

\section{Live-Demo Systems}
\label{sec:live}
We prepare a machine equipped with an RTX 3090 GPU to run our algorithm and build two capture systems to shoot live demo.
For human-scene data, our capture system consists of ten cameras positioned on a 1.6-meter beam, a piece of illumination equipment and two synchronizers, as shown in Fig.~\ref{fig:camera}(a).
In Fig.~\ref{fig:camera}(b), we position all cameras in a circle of a 2-meter radius to capture human-only data.
For live demos, capture and rendering processes are run on the same machine.
We only use the cameras with red circles, 4 cameras in Fig.~\ref{fig:camera}(a) and 6 cameras in Fig.~\ref{fig:camera}(b), as input source views.
Our method enables real-time high-quality rendering, even for challenging human-scene, human-object and multi-human interactions.

\section{Discussion}
\label{sec:conclusion}
\noindent\textbf{Conclusion.}
In this paper, we present GPS-Gaussian+, a feed-forward rendering method for both human-only data and human-scene data.
By leveraging stereo-matching and pixel-wise Gaussian parameter map regression, our method takes a significant step towards a real-time photo-realistic human-centered free-viewpoint video system from sparse views. 
When lacking depth supervision during training, a regularization term and depth residual module are designed to ensure geometry consistency and high-frequency details.
An adaptive integration, epipolar attention, is proposed in GPS-Gaussian+ to improve stereo-matching accuracy with only rendering supervision.
We demonstrate that our GPS-Gaussian+ notably improves both quantitative and qualitative results compared with baseline methods and extends original GPS-Gaussian from human-only synthesis to more scalable and general scenarios of human-centered scenes.

\noindent\textbf{Limitations.}
We notice some ghost artifacts on the white wall or the light yellow ground in the supplementary video.
This is mainly because less textured regions could increase the difficulty of stereo-matching.
Capturing more data covering more complex backgrounds to expand the diversity of the training scenes is a general solution to this issue.
To achieve this, a portable system composed of mobile phones (\textit{e.g.} Mobile-Stage dataset~\cite{xu20244k4d}) could break through the limitations of the fixed in-door capture system.
Another feasible solution is using the massive monocular videos of static scenes captured by moving cameras (\textit{e.g.} RealEstate10k~\cite{zhou2018stereo}) to pre-train the network.
However, since our method requires accurate camera calibration and strict synchronization, additional effort is required when making it practical to leverage the aforementioned data.

\bibliographystyle{unsrt}
\bibliography{IEEEabrv, main}
\end{document}